# Truncated Out-of-Bag Probability Gain: A Noise-Robust Criterion for Random-Forest Wrapper Feature Selection

**Sejong Oh**[1]

[1] Department of Software Science, Dankook University, Youngin, Republic of Korea

**Corresponding author:** Sejong Oh, sejongoh@dankook.ac.kr · ORCID: 0000-0002-9365-274X

**Acknowledgments.** NA.

## Abstract

Random-Forest wrapper feature selection can use out-of-bag (OOB) predictions as a built-in validation signal, but the criteria that decide each forward step accumulate OOB estimation noise. We propose TOP-Gain (Truncated Out-of-bag Probability Gain), which scores a candidate feature by the mean sample-level increase in the OOB probability of the true class, with decreases truncated at zero. Under a simple noise model, truncation removes about two-thirds of the noise variance contributed by unaffected samples while preserving the signal, and it makes the ranking of candidate features markedly more reproducible. In a pre-registered benchmark on 26 OpenML-CC18 datasets, TOP-Gain achieved the best mean rank in area under the ROC curve (ROC-AUC), accuracy and Macro-F1 among 15 methods; it was significantly better than 13 of 14 baselines and matched the accuracy of VSURF with 21 times less single-core computation. Within an identical wrapper, it selected significantly more stable feature sets than net-gain, Brier, log-loss and accuracy criteria without loss of accuracy. A confirmatory study on 24 further datasets replicated the advantage over all baselines. The accuracy gains were broad rather than tied to specific dataset characteristics, and, as pre-registered, the stability gain grew as relevant features became sparser.

## 1 Introduction

Selecting a small subset of informative variables is a routine step in the analysis of tabular data, from biomedical profiles to industrial and web data [1], [2]. A compact feature panel reduces measurement cost, limits overfitting and gives analysts a short list of variables to examine further [3]. Two properties matter. The panel must be *predictive*, so that a classifier trained on it generalises, and it should be *stable*, so that the same features are chosen when the data are perturbed slightly [4], [5]. We measure stability with Nogueira's estimator $\Phi$ [5], which is corrected for chance.

Feature selection methods are commonly grouped into filters, embedded methods and wrappers [1]. Filters score each feature with a statistic that does not depend on the model that will later use the features. The simplest filters are univariate tests, which rank features by how well each one separates the classes on its own. Information-theoretic filters go a step further and account for the relations between features. mRMR [6] trades the relevance of a feature to the class against its redundancy with the features already chosen, and JMI [7] rewards features that are informative jointly with them. ReliefF [8] takes a different route and scores a feature by how well it distinguishes each instance from its nearest neighbours of other classes. Embedded methods select features as a by-product of fitting a model. Elastic Net [9] drives the coefficients of uninformative features of a linear model to zero, HSIC-Lasso [10] applies the same sparsity principle to a kernel measure of dependence, and LassoNet [11] extends sparsity penalties to neural networks. A third group aggregates the results of a base selector to make them more reliable. Stability Selection [12] fits sparse models on many random subsamples and keeps the features that are selected most often, and ensemble feature selection [13] combines several selectors or several data perturbations in a similar way. Filters are fast and their rankings are typically stable, but they ignore how the features will be used and can miss features that are useful only in combination with others. Wrappers, discussed next, are usually more predictive but less stable [4].

Wrappers [14] score candidate feature subsets directly by the performance of a learning algorithm. Their scores therefore reflect predictive utility, including the non-linear effects and interactions that the classifier is able to exploit. Random Forests [15] are a natural engine for such methods: every tree is grown on a bootstrap sample, so the trees that did not see an instance provide an out-of-bag (OOB) prediction for it, and the OOB predictions of all instances estimate generalisation performance without an inner cross-validation loop. Random-Forest-based selection has accordingly become well established. The impurity and permutation importances of a single forest [15], [16] rank all features at once, although both measures are biased when features are correlated or measured on different scales [17]. RF-RFE [18] refits the forest repeatedly and removes the least important features at each round, and varSelRF [19] follows the same backward strategy but uses the OOB error to decide how many features to keep. VSURF [20] first discards features whose importance does not exceed a noise threshold and then builds nested models by forward selection. Boruta [21] compares the importance of every feature with that of randomly permuted "shadow" copies and keeps only the features that beat them. The work closest to ours is that of Velthoen et al. [22], who select variables by forward selection that minimises an OOB estimate of a proper scoring rule, the continuous ranked probability score for regression, and stop the search with a binomial test on the signs of the per-sample risk differences. We build on this RF-OOB forward-selection framework; the framework itself is not our contribution.

Our contribution concerns the criterion that each forward step maximises. The most common choice, validation accuracy, is a step function of the subset: many candidates reach exactly the same accuracy, and the tie is then broken by noise. Continuous criteria avoid such ties. Examples are the mean predicted probability of the true class, which we call the net probability gain, and the strictly proper scoring rules Brier score [23] and log-loss [24]; the latter two are the classification analogues of the criterion of [22]. All of these criteria, however, aggregate the per-sample increases and decreases of the predicted probability with equal weight. When the probabilities are OOB estimates from a modest forest, most of the per-sample changes caused by adding a single feature are estimation noise of either sign, and this noise enters the criteria in full.

Several other methods also work with per-sample quantities, but none of them truncates per-sample probability gains to select features. Conditional predictive impact [25] compares the loss of each instance under the original feature and under a knockoff-perturbed copy, averages these differences and tests them; it is designed for statistical inference about a given model rather than for subset selection. Model-X knockoffs [26] use synthetic copies of the features to select a set with a controlled false discovery rate. SHAP values [27] attribute each individual prediction to the features, and their averages are often thresholded to select features. Danyluk and Arnosti [28] score features by their effect on estimated class probabilities rather than on class labels, which makes the score sensitive to features that change the predicted probabilities without changing the predicted class.

In this paper, we propose **TOP-Gain** (Truncated Out-of-bag Probability Gain), a criterion that counts only improvements. For a candidate $f$ and the current subset $S$, TOP-Gain averages the sample-level increase $\max(0, p_i^{S\cup\{f\}} - p_i^S)$ of the OOB true-class probability and ignores the decreases. The idea of crediting a feature only for the individual cases it helps was motivated by case-based feature attribution [29], and the improvement-only form also resembles expected improvement in Bayesian optimisation [30]. The change to the RF-OOB wrapper is a single line, but it alters how the wrapper responds to estimation noise, and the rest of the paper examines this effect analytically and empirically. The main contributions are as follows.

1. **A noise-robust selection criterion with a simple explanation.** We show that truncation keeps the signal of truly improved instances while removing about two-thirds of the noise variance contributed by unaffected instances (Section 2.3). The effect can be observed directly: TOP-Gain ranks candidate features more reproducibly than every alternative criterion, and the part it discards is the noisiest component of the score (Section 4.6).
2. **Stability without an accuracy cost.** In a controlled comparison inside an identical wrapper, TOP-Gain selects more stable feature sets than the net gain, the proper scoring rules Brier and log-loss, and OOB accuracy and balanced accuracy, on 22–26 of 26 datasets each, while its accuracy is equal to or higher than theirs (Section 4.2).
3. **The accuracy of the best RF wrapper at a fraction of its cost.** Among 15 methods on 26 OpenML-CC18 datasets, TOP-Gain has the best rank in ROC-AUC, accuracy and Macro-F1. It is significantly better than 13 of 14 baselines and on par with VSURF, which needs 21 times more single-core computation (Sections 4.1, 4.4), and the advantage replicates against all 12 baselines tested on 24 independent datasets (Section 4.3).
4. **Where the stability benefit is largest, and how to tune it.** A hypothesis registered before the confirmatory data were analysed was supported: the stability gain from truncation grows as relevant features become sparser, whereas the accuracy gain is broad rather than tied to an identifiable type of dataset (Section 4.3). The size of the surrogate forest further acts as a direct handle on stability that leaves accuracy unchanged (Section 4.8).
5. **An open and reproducible implementation.** All code, raw results and the pre-registration protocol are publicly available.

The paper is organised as follows. Section 2 defines the TOP-Gain criterion, analyses why truncation reduces estimation noise and what it costs, and describes the forward-selection algorithm, its candidate pool and its computational cost. Section 3 presents the experimental design: the pre-registered dataset rules and hypotheses for the OpenML-CC18 benchmark and for the confirmatory study on the AutoML Benchmark, the compared methods, the evaluation protocol and statistics, and the runtime and sensitivity studies. Section 4 reports the results, beginning with the comparison with the baselines and the controlled comparison of criteria, followed by the confirmatory study, the computational cost, the accuracy–stability–cost trade-off, the analysis of the mechanism, ground-truth experiments and the sensitivity to the surrogate configuration. Section 5 discusses the findings, the practical use of TOP-Gain and its limitations, and Section 6 concludes. Online Resource 1 contains the dataset lists and the supplementary studies.

# 2 The TOP-Gain Criterion

## 2.1 Notation and OOB true-class probability

Let $\boldsymbol{X} \in \mathbb{R}^{N \times P}$ be the training data, with $N$ training instances (rows) and $P$ features (columns), and let $y_i \in \{1, \dots, C\}$ be the class label of instance $i$.

**OOB class probabilities.** For a feature subset $S$, we fit a Random Forest with $T$ trees on the columns $S$. Each tree is grown on a bootstrap sample of the $N$ instances. On average, a bootstrap sample leaves out about 37% of the instances, so each instance $i$ is *out of bag* (OOB) for a subset $O_i$ of about $0.37\,T$ trees (Fig. 1a). These trees did not see instance $i$ during training. Their predictions for $i$ therefore behave like predictions on new data.

For each tree $t \in O_i$, let $\pi_t(c \mid \boldsymbol{x}_i^S)$ be the class frequency of class $c$ in the leaf that instance $i$ falls into. The OOB class-probability vector of $i$ averages these frequencies over the OOB trees (Fig. 1b). We use its component for the *true* class:

$$p_i^S = \frac{1}{|O_i|} \sum_{t \in O_i} \pi_t \left( y_i \mid \boldsymbol{x}_i^S \right). \quad (1)$$

A value near 1 means that the forest on $S$ predicts instance $i$ confidently and correctly, without having seen it. An instance that is in bag for every tree receives the uniform value $1/C$, and $p_i^{\emptyset} = 1/C$.

The quantity is computed for every instance from a single forest fit, with no inner cross-validation. It is, however, an estimate from only about $0.37\,T$ trees, and its noise is the subject of Section 2.3.

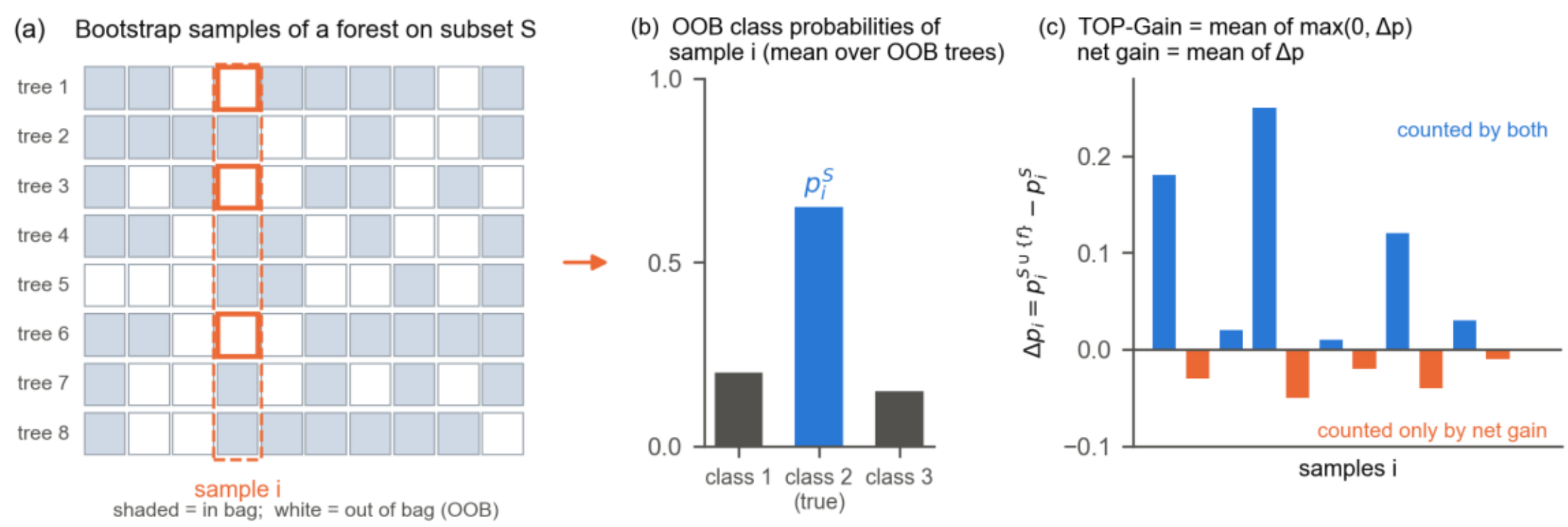

***Fig. 1*** *OOB true-class probability and the truncated gain. (a) Each tree is grown on a bootstrap sample; instance i is out of bag (white) for some of the trees. (b) Averaging the leaf class frequencies of these OOB trees gives the OOB class-probability vector of i; its true-class component is $p_i^S$. (c) Adding a candidate f changes $p_i$ by $\Delta p_i$. The net gain averages all changes; TOP-Gain averages only the increases (illustrative values)*

## 2.2 Truncated OOB probability gain

Given the current subset $S$ and a candidate $f \notin S$, TOP-Gain scores the candidate by

$$\text{TOP-Gain}(f \mid S) \;=\; \tfrac{1}{N}\textstyle\sum_{i=1}^{N} \max\left(0,\; p_i^{S\cup\{f\}} - p_i^S\right). \tag{2}$$

The net probability gain is the positive part, which is TOP-Gain, minus the negative part:

$$\tfrac{1}{N}\textstyle\sum_{i=1}^{N}\left(p_i^{S\cup\{f\}} - p_i^S\right) \;=\; \text{TOP-Gain}(f \mid S) \;-\; \tfrac{1}{N}\textstyle\sum_{i=1}^{N} \max\left(0,\; p_i^S - p_i^{S\cup\{f\}}\right). \tag{3}$$

TOP-Gain discards the negative part (Fig. 1c). Because $p_i^S$ is common to all candidates at a step, ranking by net gain is equivalent to ranking by the mean OOB true-class probability of $S \cup \{f\}$. TOP-Gain is *not* equivalent to any set-level score of $S \cup \{f\}$. It measures how many instances, and by how much, the candidate improves relative to the current subset.

## 2.3 Why truncation reduces noise

To see why truncation helps, write the observed change for instance $i$ as $\Delta_i = \delta_i + \varepsilon_i$, where $\delta_i$ is the true effect of the candidate and $\varepsilon_i \sim N(0, \sigma^2)$ is OOB estimation noise whose size depends on the number of trees for which $i$ is out of bag. At a typical forward step, a candidate leaves most instances unaffected, so that $\delta_i = 0$ and the observed change is pure noise. For such an instance the net gain receives the full noise variance $\sigma^2$, whereas TOP-Gain receives only

$$\text{Var}(\max(0, \varepsilon_i)) \;=\; \sigma^2\left(\tfrac{1}{2} - \tfrac{1}{2\pi}\right) \;\approx\; 0.34\,\sigma^2, \tag{4}$$

so truncation removes about two-thirds of the noise variance of the unaffected instances. Each of them adds a positive floor to the score,

$$E[\max(0, \varepsilon_i)] \;=\; \tfrac{\sigma}{\sqrt{2\pi}} \;\approx\; 0.40\,\sigma, \tag{5}$$

but because all candidates at a step are evaluated with the same forests and data, their noise levels are similar, the floor shifts all scores alike, and it largely cancels in the ranking. For the instances that a candidate truly improves, with $\delta_i \gg \sigma$, the truncation changes almost nothing, so the signal is kept.

A toy simulation makes the consequence concrete. With $N = 300$ instances, candidate A raises 20% of the instances by 0.10 and candidate B raises another 20% by 0.07, and every instance carries independent noise $\varepsilon_i \sim N(0, \sigma^2)$. Over 10,000 replicates per noise level, the net gain chooses the better candidate A in 99.3%, 88.4% and 77.0% of the replicates at $\sigma = 0.03$, 0.06 and 0.10, whereas TOP-Gain does so in 100.0%, 95.6% and 82.3% (Fig. 2). The gain is largest at moderate noise. This is the regime of a 40-tree surrogate, with about 15 OOB trees per instance, in the middle of the search; Section 4.6 examines whether this prediction holds on real data.

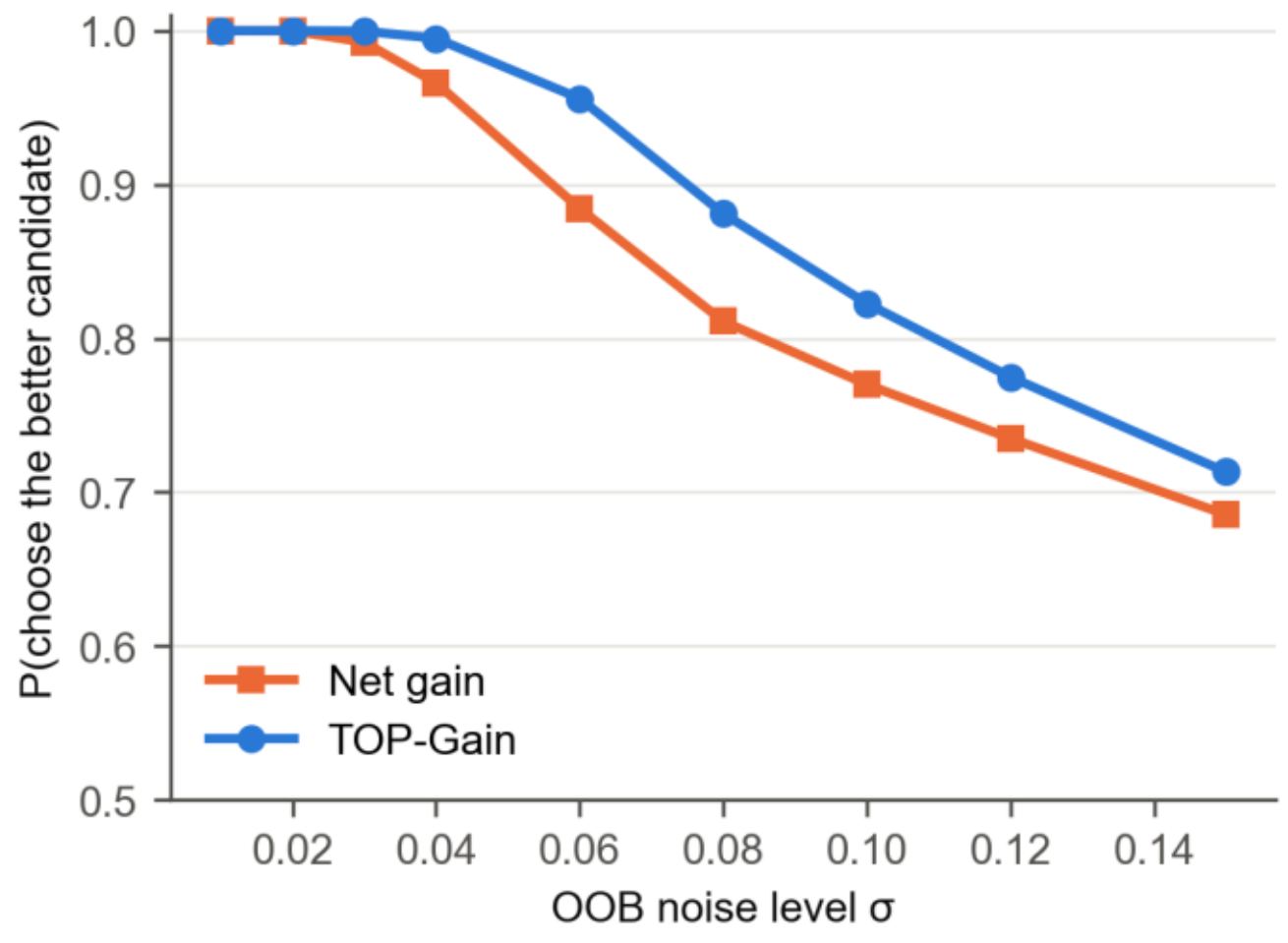

***Fig. 2*** *Toy simulation of Section 2.3: probability of choosing the truly better candidate under the net gain and under TOP-Gain, as a function of the noise level σ (10,000 replicates per level)*

The argument assumes that negative changes are mostly noise, and TOP-Gain indeed does not penalise real harm. The argument rests on the assumption that negative changes are mostly noise rather than real harm. We discuss when this assumption may fail in Section 5.

## 2.4 Forward selection algorithm

Algorithm 1 embeds the criterion in a standard forward search.

**Algorithm 1.** TOP-Gain forward selection

```
Input : training data (X, y), target size K, surrogate size T, seed s
Output: ordered feature list S

1  A  <- CandidatePool(X, y, K)           # all features if P <= 150, else Section 2.5
2  S  <- [ ];  p_base <- uniform 1/C
3  for step = 1 .. K:
4      for each f in A \ S  (in parallel):
5          p_f      <- OOB true-class probabilities of an RF(T trees, seed s) on S ∪ {f}
6          score(f) <- mean_i max(0, p_f[i] - p_base[i])
7      f* <- argmax score(f)   (ties broken by feature name)
8      S  <- S + [f*]
9      p_base <- OOB true-class probabilities of an RF(T trees, seed s + step) on S
10 return S
```

Because the output is an ordered list, one run with $K_{\max}$ steps yields nested panels for every smaller $K$. All candidates at a step are evaluated with the same seed, so their score differences are not caused by different bootstrap draws.

Every experiment in Sections 3 and 4 uses surrogate forests with $T = 40$ trees, a maximum depth of 5 and a minimum leaf size of 2. These values were fixed a priori and not tuned on the benchmark data. The number of trees governs a trade-off between cost and noise: each candidate at each step needs its own forest, so $T$ multiplies the total cost, while with $T = 40$ every instance is out of bag for about 15 trees, which gives a usable probability estimate at a cost of seconds per fold. Shallow trees keep each fit cheap and the OOB probabilities smooth. Section 4.8 shows that accuracy is insensitive to $T$ between 20 and 160 and to the depth, while stability increases with $T$. After each step the reference probabilities are refitted with seed $s + \text{step}$ (line 9); a variant that reuses seed $s$ gives equivalent results (Online Resource 1, Section S4).

## 2.5 Candidate pool for higher-dimensional data

When $P \leq 150$, every feature is a candidate at every step. When $P > 150$, evaluating all features at every step would be wasteful, and the search is restricted to a candidate pool $A$ that is chosen once, on the training fold, before the first step. Let $m = \min(P, \max(200, 5K))$, and let three screening lists $L_{\text{RF}}$, $L_F$ and $L_{\text{var}}$ contain the $\lfloor m/2 \rfloor$ top-ranked features by the impurity importance of a 40-tree Random Forest on all features, by the ANOVA $F$ statistic and by the variance, respectively. The pool is the union of the three lists,

$$A = L_{\text{RF}} \cup L_F \cup L_{\text{var}}, \qquad \lfloor m/2 \rfloor \leq |A| \leq 3\lfloor m/2 \rfloor, \tag{6}$$

so a feature enters the pool if any one of the three criteria ranks it highly. The lists are complementary: the forest importance captures non-linear and interaction effects, the $F$ statistic captures marginal class separation, and the variance retains high-variance features that the other two criteria may miss in small samples. The size of the union depends on how much the lists overlap; with $K \leq 20$, the pool contains between 100 and 300 features.

## 2.6 Computational cost

A run fits $O(K \cdot |A|)$ forests of $T$ shallow trees on at most $K$ features. For $P > 150$ the pool size $|A|$ is bounded independently of $P$, so the cost grows only weakly with $P$. The candidate evaluations within a step are independent and parallelise across cores. Measured selection times are reported in Section 4.4.

# 3 Experimental Design

## 3.1 Benchmark suites and pre-registration

To support claims that hold beyond a hand-picked set of datasets, the dataset rules, the dataset characteristics (moderators) and the hypotheses were written down in a protocol before any selector was run on the new data. Every later change was appended as a dated amendment together with its reason, and the protocol and its amendments are public (Online Resource 1, Section S6). Table 1 summarises the two benchmark suites that resulted.

**Table 1** The two benchmark suites. $N$ and $P$ are the numbers of instances and features in use, after subsampling to at most 1,500 instances and one-hot encoding

| | OpenML-CC18 (main benchmark) | AutoML Benchmark (confirmatory study) |
|---|---|---|
| Source | OpenML-CC18, 72 curated tasks [31] | OpenML suite 271 [32], minus every CC18 dataset |
| Eligibility | R1: $P \geq 20$; R2: $N \leq 50{,}000$; R3: $0.60 \leq$ AUC of a full-feature RF $< 0.99$ | R1 and R3 unchanged; R2 relaxed to $N \leq 10^6$ |
| Eligible datasets | 26 | 24 |
| $N$ in use | 522–1,500 | 100–1,500 |
| $P$ in use | 18–1,776 | 20–71,640 |
| Classes | 2–10 | 2–50 |
| Majority/minority ratio | 1.0–15.3 | 1.0–104.4 |
| Relevance density M-rel | 0.03–1.00 | 0.01–0.95 |
| Methods | TOP-Gain, 14 baselines, 5 alternative criteria | TOP-Gain, net gain, 12 Python baselines |
| Repeats of 5-fold CV | 3 | 3 |
| Hypotheses | H1–H3 (pre-registered) | H4a, H4b (registered after the CC18 analysis) |

The main benchmark draws on OpenML-CC18, a suite of 72 curated classification tasks [31]. A dataset is eligible if it satisfies three structural rules:

- **R1:** $P \geq 20$;
- **R2:** $N \leq 50{,}000$;
- **R3:** $0.60 \leq$ ROC-AUC of a full-feature reference Random Forest $< 0.99$.

The first rule excludes problems where selection is nearly trivial, the second keeps the computation feasible, and the third excludes datasets at the ceiling or near chance level, on which all selectors perform alike. Twenty-six datasets are eligible (Table 1; Online Resource 1, Table S1).

Before any selector was run, we also computed moderators that describe each dataset. Non-linearity M-nl is the ROC-AUC of the full-feature Random Forest minus that of an L2-regularised logistic regression. Redundancy M-red is the mean, over features, of the maximum absolute Spearman correlation with any other feature. Headroom M-head is one minus the ROC-AUC of the full-feature Random Forest. Relevance density M-rel is the fraction of features whose ANOVA $F$ statistic is significant at a false discovery rate of 0.05. Dimensionality, class imbalance and the share of categorical features complete the set.

The pre-registered hypotheses concern two per-dataset outcomes: ΔAUC, the ROC-AUC of TOP-Gain minus the median ROC-AUC of the baselines, and ΔΦ, the stability of TOP-Gain minus that of the net-gain criterion. They were derived from the noise mechanism, which suggested that truncation helps most when many candidates offer similar gains:

- **H1:** ΔAUC increases with non-linearity M-nl;
- **H2:** ΔΦ increases with redundancy M-red;
- **H3:** ΔAUC decreases as headroom M-head shrinks.

Each hypothesis is tested by a Spearman permutation test, with Holm's correction across the three.

After the CC18 analysis had been completed, H1–H3 were recorded as not supported (Section 4.3). A new hypothesis was then registered from the exploratory CC18 results, together with a new and independent pool of datasets, before any result on that pool was computed:

- **H4a (primary):** ΔAUC decreases with relevance density M-rel;

- **H4b (secondary):** ΔΦ decreases with M-rel.

The confirmatory pool is OpenML suite 271, the AutoML Benchmark (AMLB) collection of classification tasks [32], from which every CC18 dataset was removed. Rules R1 and R3 were kept unchanged, and R2 was relaxed to $N \leq 10^6$, because all data are subsampled anyway; 24 datasets are eligible (Table 1). Both hypotheses are tested by one-sided Spearman permutation tests with 10,000 permutations and the alternative $\rho < 0$ at $\alpha = 0.05$, and H4a counts as confirmed only if its test rejects.

## 3.2 Compared methods

TOP-Gain is compared with 14 established selectors. The filters are the ANOVA $F$ test, mRMR [6], JMI [7] and ReliefF [8]. Elastic Net [9], HSIC-Lasso [10] and Stability Selection [12] represent embedded and aggregation methods. The model-based methods are Boruta [21], SVM-RFE [33], the impurity and permutation importances of a Random Forest [16] and RF-RFE [18], and we include the original R implementations of varSelRF [19] and VSURF [20] with their package defaults. Every selector returns a ranked list, of which the first $K$ features are evaluated; for varSelRF and VSURF, which return sets, the ranking follows their elimination order and importances. As registered, varSelRF and VSURF were not run on the AMLB pool for computational reasons.

To isolate the effect of truncation, we also replace the TOP-Gain criterion by five alternatives inside exactly the same forward wrapper, with the same candidate pool, forests, seeds and tie-breaking. The alternatives are the net probability gain, the OOB Brier score, the OOB log-loss with probabilities clipped at $10^{-3}$, OOB accuracy and OOB balanced accuracy.

## 3.3 Protocol

Each dataset is evaluated by stratified 5-fold cross-validation, repeated three times, where repeat $r \in \{1,2,3\}$ uses seed $r$ both for the folds and for every selector. Datasets with more than 1,500 instances are first reduced to 1,500 by stratified subsampling; categorical variables are one-hot encoded and numeric missing values are imputed with training-fold medians. The panel sizes are $K \in \{5,10,15,20\}$, keeping only the values with $K \leq P/2$. Each panel is evaluated with four downstream classifiers: a Random Forest (100 trees, balanced class weights), XGBoost (100 rounds, depth 4), an RBF-kernel support vector machine and an L2-regularised logistic regression. A selector call that failed for an infrastructure reason, such as lack of memory or a failure of parallel transport, was re-run once in full, and no result was ever replaced by that of another method.

The primary outcomes were fixed before the analysis: ROC-AUC (one-vs-rest macro average) for predictive performance and Nogueira's Φ over the five folds for stability. Accuracy, Macro-F1 and selection time are secondary outcomes.

## 3.4 Statistical analysis

Following Demšar [34], the unit of analysis is the dataset. For each dataset and method, each metric is averaged over the $K$ values, the four classifiers and the three repeats; the panels for different $K$ are nested prefixes of one ordered list and therefore not independent units. For each metric we apply a Friedman test with the Nemenyi critical difference (CD) at $\alpha = 0.05$. For TOP-Gain versus each other method we report the mean paired difference with a bootstrap 95% confidence interval (CI; 10,000 resamples), the wins and losses over datasets, and the two-sided Wilcoxon signed-rank $p$-value, Holm-corrected within each family of comparisons (the 14 baselines or the five criteria).

## 3.5 Runtime study

The benchmark ran each selector with its own default parallelism. TOP-Gain and the Python baselines used ten cores, VSURF used a cluster of nine workers, and varSelRF is single-threaded, so the benchmark timings are not directly comparable. We therefore re-timed all 15 methods under two stated settings. In the first, every method runs on one core. In the second, every method that supports parallelism uses ten cores; varSelRF has no parallel option.

The measurements use the same fold (repeat 1, fold 1) of all 26 CC18 datasets and run one call at a time on an otherwise idle machine. A discarded warm-up call per method absorbs the start-up of the worker processes and of Rscript. We record the wall-clock time of each selector call and, for varSelRF and VSURF, also the R-internal fitting time, and we compare the selected lists of the two settings. The machine has an Intel Core Ultra 5 225 processor with ten cores and 15.4 GB of RAM and runs Windows 11. The software comprises Python 3.14.3 with scikit-learn 1.8.0, NumPy 2.4.3, pandas 3.0.1 and joblib 1.5.3, and R 4.6.1 with randomForest 4.7-1.2, varSelRF 0.7-9 and VSURF 1.2.1.

## 3.6 Sensitivity study and further studies

A further study, also registered before it was run, examines the sensitivity of TOP-Gain to its surrogate configuration by changing one setting at a time from the default. On all 26 CC18 datasets with three repeats, the number of trees is set to $T \in \{20,80,160\}$ and the maximum depth to 3 or 8. On the 16 AMLB datasets with $P > 150$, where the candidate pool is active, the pool size is scaled by 0.5 or 2 with one repeat. Each variant is compared with default TOP-Gain by the dataset-level statistics of Section 3.4, Holm-corrected over the variants.

Online Resource 1 reports three further studies. The first is the original ten-dataset biomedical benchmark, which includes microarray and proteomics data with $P \gg N$ (Section S2). The second comprises ground-truth and controlled-imbalance studies on synthetic data and on OVA_Lung transcriptomic data (Section S3). The third is a common-seed robustness check and an ablation of multi-objective extensions (Section S4).

## 3.7 Use of generative AI tools

An AI assistant, Claude (Anthropic), was used to implement the experiment and analysis code, to run the experiments and statistical analyses, and to draft parts of the manuscript text. The authors conceived the method, designed the study, wrote the pre-registration protocol and decided every amendment, interpreted the results, and reviewed and verified all code, reported numbers, references and text. Every number in the paper is produced by the public analysis scripts from the stored result files. No generative AI was used to create images: the figures were drawn with plotting code from the experimental data, and the schematic Fig. 1 uses illustrative values. The authors take full responsibility for the content of the article.

# 4 Results

## 4.1 Main benchmark: TOP-Gain versus 14 baselines (CC18)

Table 2 and Fig. 3 summarise the comparison with the 14 baselines on the 26 CC18 datasets.

**Table 2** CC18 benchmark (26 datasets × 3 repeats). Values are means over datasets of the dataset-level means. Rank is the Friedman mean rank of ROC-AUC (15 methods; CD = 4.21). ΔAUC is TOP-Gain minus the baseline, with a bootstrap 95% CI; W/L counts datasets; $p_{\text{Holm}}$ is the Holm-adjusted Wilcoxon $p$ for ROC-AUC (14 comparisons). Time is the median over datasets of the selection time per fold under the benchmark's default parallelism (Section 4.4 gives controlled timings)

| Method | Rank (AUC) ↓ | ROC-AUC | Accuracy | Macro-F1 | Φ | ΔAUC [95% CI] | W/L | $p_{\text{Holm}}$ | Time/fold (s) |
|---|---|---|---|---|---|---|---|---|---|
| **TOP-Gain** | **3.92** | **0.8692** | **0.8200** | **0.7285** | 0.514 | – | – | – | 3.8 |
| VSURF | 4.15 | 0.8663 | 0.8178 | 0.7248 | 0.722 | 0.003 [0.000, 0.006] | 13/13 | 0.27 | 18.3 |
| RF-RFE | 5.88 | 0.8603 | 0.8124 | 0.7170 | 0.782 | 0.009 [0.002, 0.020] | 18/8 | 0.028 | 1.4 |
| HSIC-Lasso | 6.50 | 0.8595 | 0.8090 | 0.7141 | 0.694 | 0.010 [0.002, 0.022] | 19/7 | 0.015 | 0.15 |
| varSelRF | 6.62 | 0.8603 | 0.8139 | 0.7177 | 0.847 | 0.009 [0.004, 0.015] | 20/6 | 0.015 | 31.7 |
| RF impurity | 6.69 | 0.8620 | 0.8119 | 0.7172 | **0.853** | 0.007 [0.003, 0.013] | 18/8 | 0.028 | 0.5 |
| RF permutation | 6.96 | 0.8619 | 0.8134 | 0.7181 | 0.518 | 0.007 [0.004, 0.012] | 23/3 | 0.001 | 4.5 |
| Elastic Net | 7.77 | 0.8560 | 0.8061 | 0.7108 | 0.663 | 0.013 [0.004, 0.028] | 20/6 | 0.004 | 0.05 |
| Stability Selection | 8.69 | 0.8470 | 0.7936 | 0.6998 | 0.682 | 0.022 [0.006, 0.051] | 20/6 | 0.005 | 0.05 |
| ANOVA | 8.77 | 0.8534 | 0.8032 | 0.7070 | 0.844 | 0.016 [0.008, 0.025] | 23/3 | 0.001 | 0.001 |
| JMI | 9.00 | 0.8457 | 0.7999 | 0.6994 | 0.549 | 0.024 [0.011, 0.044] | 22/4 | <0.001 | 0.29 |
| Boruta | 9.96 | 0.8445 | 0.7916 | 0.6926 | 0.833 | 0.025 [0.014, 0.037] | 22/4 | <0.001 | 6.8 |
| mRMR | 10.04 | 0.8409 | 0.7976 | 0.6952 | 0.551 | 0.028 [0.014, 0.050] | 22/4 | <0.001 | 0.28 |
| SVM-RFE | 11.65 | 0.8404 | 0.8019 | 0.6993 | 0.549 | 0.029 [0.016, 0.045] | 24/2 | <0.001 | 0.15 |
| ReliefF | 13.38 | 0.8016 | 0.7787 | 0.6609 | 0.455 | 0.068 [0.036, 0.107] | 24/2 | <0.001 | 0.05 |

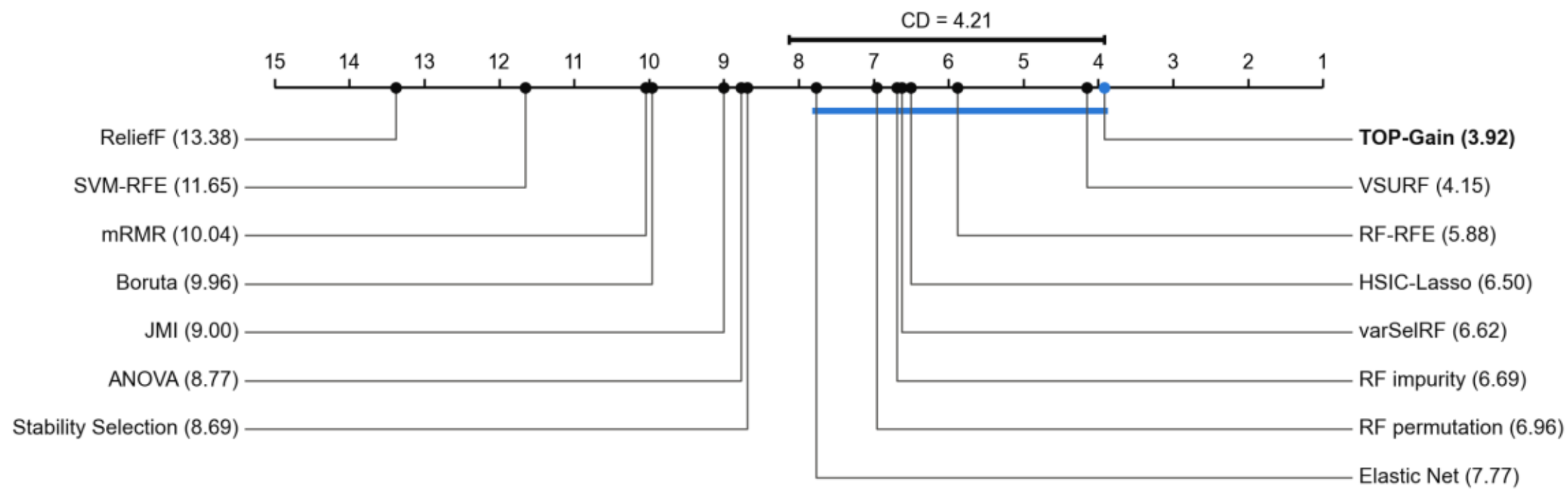

**Fig. 3** Critical-difference diagram of the Friedman mean ranks of ROC-AUC on the 26 CC18 datasets (15 methods, three repeats; lower rank is better). The blue bar below the axis spans the methods whose mean rank lies within one critical difference (CD) of TOP-Gain, that is, the methods not significantly different from it (Nemenyi test, α = 0.05)

The Friedman test rejects equal performance for all three predictive metrics ($p < 10^{-13}$). TOP-Gain has the best mean rank for ROC-AUC (3.92), accuracy (3.54) and Macro-F1 (3.50), and VSURF is second in all three (4.15, 4.44 and 4.54). In ROC-AUC, TOP-Gain is significantly better than 13 of the 14 baselines ($p_{\mathrm{Holm}} \leq 0.028$), with mean differences between 0.007 and 0.068, and the same holds for accuracy ($p_{\mathrm{Holm}} \leq 0.016$) and Macro-F1 ($p_{\mathrm{Holm}} \leq 0.006$). The exception is VSURF. Its ROC-AUC differs from that of TOP-Gain by only +0.003 (CI [0.000, 0.006]; 13 wins and 13 losses; $p_{\mathrm{Holm}} = 0.27$), and although the accuracy (+0.002, $p_{\mathrm{Holm}} = 0.029$) and Macro-F1 (+0.004, $p_{\mathrm{Holm}} = 0.044$) differences are significant, they are very small. We therefore describe TOP-Gain as on par with VSURF and ahead of all other baselines.

The ranking also holds for each downstream classifier separately (Online Resource 1, Table S5). TOP-Gain has the best ROC-AUC rank with the Random Forest and XGBoost classifiers and is second, behind VSURF by less than 0.4 rank, with the support vector machine and logistic regression; in accuracy and Macro-F1 it ranks first for all four classifiers.

Stability shows the opposite picture. With $\Phi = 0.514$, TOP-Gain has the lowest mean stability rank of the 15 methods. RF impurity (0.853), varSelRF (0.847), ANOVA (0.844) and Boruta (0.833) are the most stable, and TOP-Gain is comparable to RF permutation importance, mRMR, JMI, SVM-RFE and ReliefF ($p_{\mathrm{Holm}} = 1.0$). This is the familiar trade-off between the stability of filters and importance rankings and the accuracy of wrappers [4]. Among the wrapper criteria, however, TOP-Gain is the most stable (Section 4.2), and on the AMLB pool its absolute stability is in the middle of the ranking (Section 4.3).

## 4.2 The effect of truncation: criterion comparison (CC18)

Because only the criterion changes across the rows of Table 3, this comparison isolates the effect of truncation.

**Table 3** Six criteria inside the same RF-OOB forward wrapper (CC18; Friedman mean ranks among the six criteria, CD = 1.48). Δ is TOP-Gain minus the alternative; $p_{\mathrm{Holm}}$ is over five comparisons.

| Criterion | Rank AUC ↓ | ROC-AUC | Macro-F1 | Rank Φ ↓ | Φ | ΔAUC (W/L, $p_{\mathrm{Holm}}$) | ΔΦ [95% CI] (W/L, $p_{\mathrm{Holm}}$) |
|---|---|---|---|---|---|---|---|
| **TOP-Gain** | **2.23** | **0.8692** | **0.7285** | **1.42** | **0.514** | – | – |
| OOB log-loss | 2.35 | 0.8682 | 0.7277 | 3.15 | 0.417 | +0.001 (13/13, 0.41) | +0.097 [0.060, 0.134] (22/4, <0.001) |
| OOB Brier score | 3.06 | 0.8664 | 0.7279 | 3.10 | 0.420 | +0.003 (18/8, 0.016) | +0.094 [0.062, 0.127] (24/2, <0.001) |
| Net probability gain | 3.73 | 0.8655 | 0.7256 | 2.40 | 0.449 | +0.004 (21/5, <0.001) | +0.065 [0.037, 0.093] (22/4, <0.001) |
| OOB balanced accuracy | 4.62 | 0.8589 | 0.7174 | 5.38 | 0.276 | +0.010 (22/4, <0.001) | +0.238 [0.198, 0.274] (25/1, <0.001) |
| OOB accuracy | 5.02 | 0.8504 | 0.7127 | 5.54 | 0.265 | +0.019 (24/2, <0.001) | +0.249 [0.210, 0.287] (26/0, <0.001) |

Stability separates the criteria clearly. Against every alternative, TOP-Gain is more stable on 22–26 of the 26 datasets ($p_{\mathrm{Holm}} < 0.001$); its Φ exceeds that of the proper scoring rules by about 0.095, a relative gain of 23%, and that of the discrete criteria by about 0.24, a relative gain of 90%. Predictive performance does not suffer: TOP-Gain's ROC-AUC equals that of the log-loss criterion (13 wins and 13 losses) and is significantly higher than that of the other four criteria. The comparison with the net gain is the most direct one, because the two criteria differ only in whether sample-level decreases are counted. Discarding them raises Φ by 0.065 (22 wins, 4 losses) and ROC-AUC by 0.004 (21 wins, 5 losses). Truncation therefore buys stability without any loss of accuracy.

## 4.3 Confirmatory study (AMLB) and the pre-registered hypotheses

On the 24 AMLB datasets, none of which was used to develop TOP-Gain or its hypotheses, the advantage over the baselines replicates and is, if anything, stronger (Table 4). TOP-Gain has the best Friedman rank among the 13 methods for ROC-AUC (2.67), accuracy (3.13) and Macro-F1 (3.21), with RF-RFE second in all three (4.21, 4.79 and 5.08). It is significantly better than all 12 baselines on all three metrics ($p_{\text{Holm}} \leq 0.029$); even against the closest one, RF-RFE, the ROC-AUC difference is +0.011 (CI [0.004, 0.019]; 17 wins, 7 losses; $p_{\text{Holm}} = 0.011$). The comparison with the net gain also replicates: TOP-Gain again selects more stable sets (ΔΦ = +0.080, CI [0.046, 0.115]; 22 wins, 2 losses; $p_{\text{Holm}} < 0.001$) and has a slightly higher ROC-AUC (+0.006; 18 wins, 6 losses; $p_{\text{Holm}} = 0.011$), accuracy and Macro-F1 (+0.002 each, $p_{\text{Holm}} = 0.036$). Its absolute stability ranks seventh of 13 on this pool; it is significantly below RF impurity, ANOVA, RF-RFE and Boruta and not significantly different from the other eight baselines.

**Table 4** Confirmatory AMLB study (24 datasets × 3 repeats; 12 Python baselines). Rank = Friedman mean rank (13 methods, CD = 3.72).

| Method | Rank AUC ↓ | ROC-AUC | Rank Φ ↓ | Φ | ΔAUC [95% CI] | W/L | $p_{\text{Holm}}$ |
|---|---|---|---|---|---|---|---|
| **TOP-Gain** | **2.67** | **0.8069** | 8.21 | 0.444 | – | – | – |
| RF-RFE | 4.21 | 0.7958 | 4.50 | 0.659 | 0.011 [0.004, 0.019] | 17/7 | 0.011 |
| RF impurity | 5.12 | 0.7925 | **2.50** | **0.739** | 0.014 [0.007, 0.023] | 19/5 | 0.005 |
| RF permutation | 5.54 | 0.7882 | 9.17 | 0.386 | 0.019 [0.009, 0.030] | 21/3 | <0.001 |
| HSIC-Lasso | 6.58 | 0.7833 | 6.33 | 0.524 | 0.024 [0.014, 0.035] | 22/2 | <0.001 |
| Boruta | 6.75 | 0.7751 | 4.21 | 0.655 | 0.032 [0.015, 0.053] | 19/5 | <0.001 |
| JMI | 7.21 | 0.7781 | 9.33 | 0.336 | 0.029 [0.013, 0.049] | 19/5 | <0.001 |
| mRMR | 7.96 | 0.7711 | 10.29 | 0.313 | 0.036 [0.018, 0.058] | 21/3 | <0.001 |
| ANOVA | 8.04 | 0.7593 | 2.58 | 0.735 | 0.048 [0.026, 0.077] | 21/3 | <0.001 |
| Elastic Net | 8.67 | 0.7597 | 6.71 | 0.478 | 0.047 [0.031, 0.065] | 21/3 | <0.001 |
| SVM-RFE | 8.92 | 0.7656 | 9.29 | 0.356 | 0.041 [0.029, 0.054] | 23/1 | <0.001 |
| Stability Selection | 9.12 | 0.7509 | 8.42 | 0.412 | 0.056 [0.033, 0.085] | 22/2 | <0.001 |
| ReliefF | 10.21 | 0.7349 | 9.46 | 0.360 | 0.072 [0.052, 0.093] | 23/1 | <0.001 |
| *Net probability gain* | – | 0.8006 | – | 0.364 | 0.006 [0.002, 0.011] | 18/6 | 0.011 |

Table 5 reports the tests of all five pre-registered hypotheses.

**Table 5** Tests of the pre-registered hypotheses. Spearman correlation between the moderator and the per-dataset outcome, with a permutation $p$-value.

| Hypothesis | Data | Moderator → outcome | Predicted | ρ | $p$ | Result |
|---|---|---|---|---|---|---|
| H1 | CC18 (26) | M-nl → ΔAUC | ρ > 0 | +0.37 | 0.065 (Holm 0.20) | not supported |
| H2 | CC18 (26) | M-red → ΔΦ | ρ > 0 | −0.29 | 0.15 (Holm 0.29) | not supported (opposite sign) |
| H3 | CC18 (26) | M-head → ΔAUC | ρ > 0 | −0.12 | 0.55 (Holm 0.55) | not supported |
| **H4a** (primary) | AMLB (24) | M-rel → ΔAUC | ρ < 0 | −0.28 | 0.089 (one-sided) | **not confirmed** |
| **H4b** (secondary) | AMLB (24) | M-rel → ΔΦ | ρ < 0 | **−0.57** | **0.004** (one-sided) | **supported** |

No moderator of the accuracy gain was confirmed. Hypotheses H1–H3 were not supported on CC18, and H4a, the accuracy hypothesis suggested by the CC18 data, was not confirmed on the new data. TOP-Gain's accuracy advantage thus appears to be broad: ΔAUC is positive on 24 of the 26 CC18 datasets and on all 24 AMLB datasets, and it is not concentrated in a structurally identifiable subgroup that we could have predicted.

The stability moderator, in contrast, was supported. The stability gain from truncation grows as the relevant features become sparser (Fig. 4), which replicates the exploratory CC18 correlation (ρ = −0.42) on independent data. For example, ΔΦ is +0.27 on porto-seguro, +0.28 on kick and +0.19 on KDDCup09_appetency, all with $M_{\text{rel}} \leq 0.025$, but at most 0.01 on covertype, dilbert, robert and volkert, all with $M_{\text{rel}} \geq 0.71$. This is what the noise argument of Section 2.3 predicts: when few features are relevant, most candidates at each step are uninformative and offer similar, noise-dominated gains, and this is the regime in which removing noise from the score matters most.

An exploratory analysis, which is post hoc and unadjusted, further showed that on AMLB the accuracy gain increased with non-linearity M-nl (ρ = +0.50, $p = 0.011$) and with $\log_{10}P$ (ρ = +0.52). The direction for M-nl agrees with the unconfirmed CC18 hypothesis H1 (ρ = +0.37). We report these as observations to be tested in future work, not as findings.

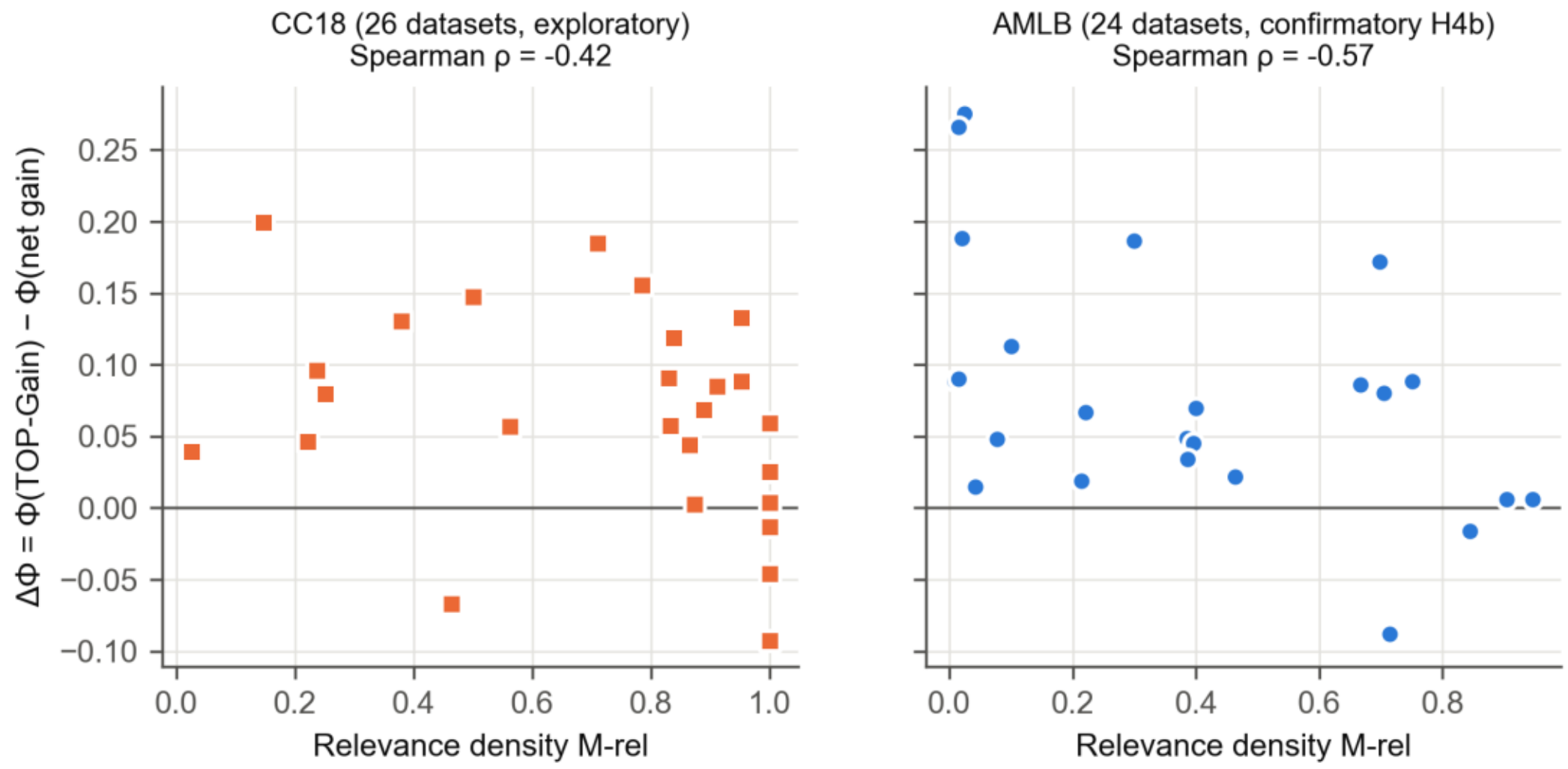

*Fig. 4 Relevance density and the stability gain from truncation: ΔΦ = Φ(TOP-Gain) − Φ(net gain) per dataset against the relevance density M-rel, on CC18 (exploratory) and on the AMLB pool (confirmatory test of H4b)*

## 4.4 Computational cost under identical settings

Table 6 reports the selection time of every method on one fold of each CC18 dataset under the two settings of Section 3.5.

**Table 6** Selection time under identical settings (26 CC18 datasets, repeat 1, fold 1; one call per method and dataset). Total = sum over datasets; ratio = total time relative to TOP-Gain; speed-up = total time on one core divided by that on ten cores. varSelRF has no parallel option.

| Method | Total, 1 core (s) | Median, 1 core (s) | Ratio to TOP-Gain, 1 core | Total, 10 cores (s) | Ratio to TOP-Gain, 10 cores | Speed-up 10 vs 1 |
|---|---|---|---|---|---|---|
| **TOP-Gain** | 1,075 | 17 | 1.00 | 165 | 1.00 | 6.5 |
| VSURF | 22,859 | 61 | 21.27 | 4,736 | 28.72 | 4.8 |
| varSelRF | 2,556 | 24 | 2.38 | 2,609 | 15.82 | 1.0 |
| RF permutation | 701 | 5.2 | 0.65 | 382 | 2.31 | 1.8 |
| Boruta | 206 | 4.7 | 0.19 | 125 | 0.76 | 1.6 |
| RF-RFE | 37 | 1.1 | 0.03 | 25 | 0.15 | 1.5 |
| SVM-RFE | 37 | 0.1 | 0.03 | 37 | 0.22 | 1.0 |
| RF impurity | 23 | 0.7 | 0.02 | 10 | 0.06 | 2.2 |

| Method | Total, 1 core (s) | Median, 1 core (s) | Ratio to TOP-Gain, 1 core | Total, 10 cores (s) | Ratio to TOP-Gain, 10 cores | Speed-up 10 vs 1 |
|---|---|---|---|---|---|---|
| mRMR | 22 | 0.3 | 0.02 | 21 | 0.13 | 1.1 |
| JMI | 18 | 0.3 | 0.02 | 17 | 0.10 | 1.1 |
| Stability Selection | 16 | 0.04 | 0.01 | 15 | 0.09 | 1.0 |
| HSIC-Lasso | 15 | 0.2 | 0.01 | 7.4 | 0.04 | 2.1 |
| Elastic Net | 3.7 | 0.05 | 0.00 | 3.7 | 0.02 | 1.0 |
| ReliefF | 0.5 | 0.01 | 0.00 | 0.6 | 0.00 | 0.8 |
| ANOVA | 0.07 | 0.00 | 0.00 | 0.07 | 0.00 | 0.9 |

VSURF, the only baseline as accurate as TOP-Gain, is the most expensive method in both settings. On one core it needs 21 times the total time of TOP-Gain (22,859 s versus 1,075 s), and on ten cores 29 times. TOP-Gain is faster on 25 of the 26 datasets on one core and on all 26 on ten cores, with median per-dataset ratios of 3.0 and 4.1. The gap grows with dimensionality: on Bioresponse ($P = 1{,}776$), VSURF needs 17,798 s on one core against 162 s for TOP-Gain, a factor of 110, and on Internet-Advertisements ($P = 1{,}558$) 3,139 s against 119 s.

The comparison with varSelRF depends on the setting. On one core the two methods cost about the same: varSelRF needs 2.4 times the total time of TOP-Gain, with a median per-dataset ratio of 1.3, and TOP-Gain is faster on 16 of the 26 datasets. With ten cores the ratio rises to 16, because TOP-Gain parallelises well whereas varSelRF does not. In fact, TOP-Gain has the largest parallel speed-up of all methods, 6.5 on ten cores against 4.8 for VSURF, since its candidate evaluations within a step are independent. The filters, the embedded methods and the single-fit importance rankings are much cheaper than every wrapper; Boruta and RF permutation importance cost about a fifth and two-thirds of TOP-Gain on one core.

Three further observations support these comparisons. For varSelRF and VSURF, the R-internal fitting time accounts for more than 99.7% of the measured wall-clock time on one core (2,549 of 2,556 s and 22,852 of 22,859 s), so the transfer of the data to R does not affect the results. The worker count did not change any TOP-Gain selection, which was identical under both settings on all 26 datasets, whereas VSURF's selections differed on 24 of the 26 datasets because its parallel and sequential modes draw different random-number streams; the accuracy results of Section 4.1 refer to its parallel mode. Finally, on the AMLB pool, TOP-Gain's selection time grew only slowly with $P$, from 2 s per fold at $P = 20$ to 65 s at $P = 71{,}640$, because the candidate pool is bounded (Section 2.5); RF permutation importance, by contrast, needed about 10,000 s per fold at $P = 71{,}640$.

## 4.5 Accuracy, stability and cost

Fig. 5 combines the three aspects by placing every method according to its accuracy rank and stability rank on CC18, with a marker size that grows with its single-core selection time. Three groups emerge. The filters and importance rankings, such as ANOVA, RF impurity and Boruta, are stable and fast but less accurate. VSURF is accurate and stable but by far the most expensive method, while varSelRF is stable and moderately accurate at a single-core cost similar to that of TOP-Gain. TOP-Gain is the most accurate method at a moderate cost, with lower absolute stability. No method dominates TOP-Gain in accuracy: all methods that are more stable are less accurate, and the only method that is equally accurate, VSURF, needs 21 times more single-core computation.

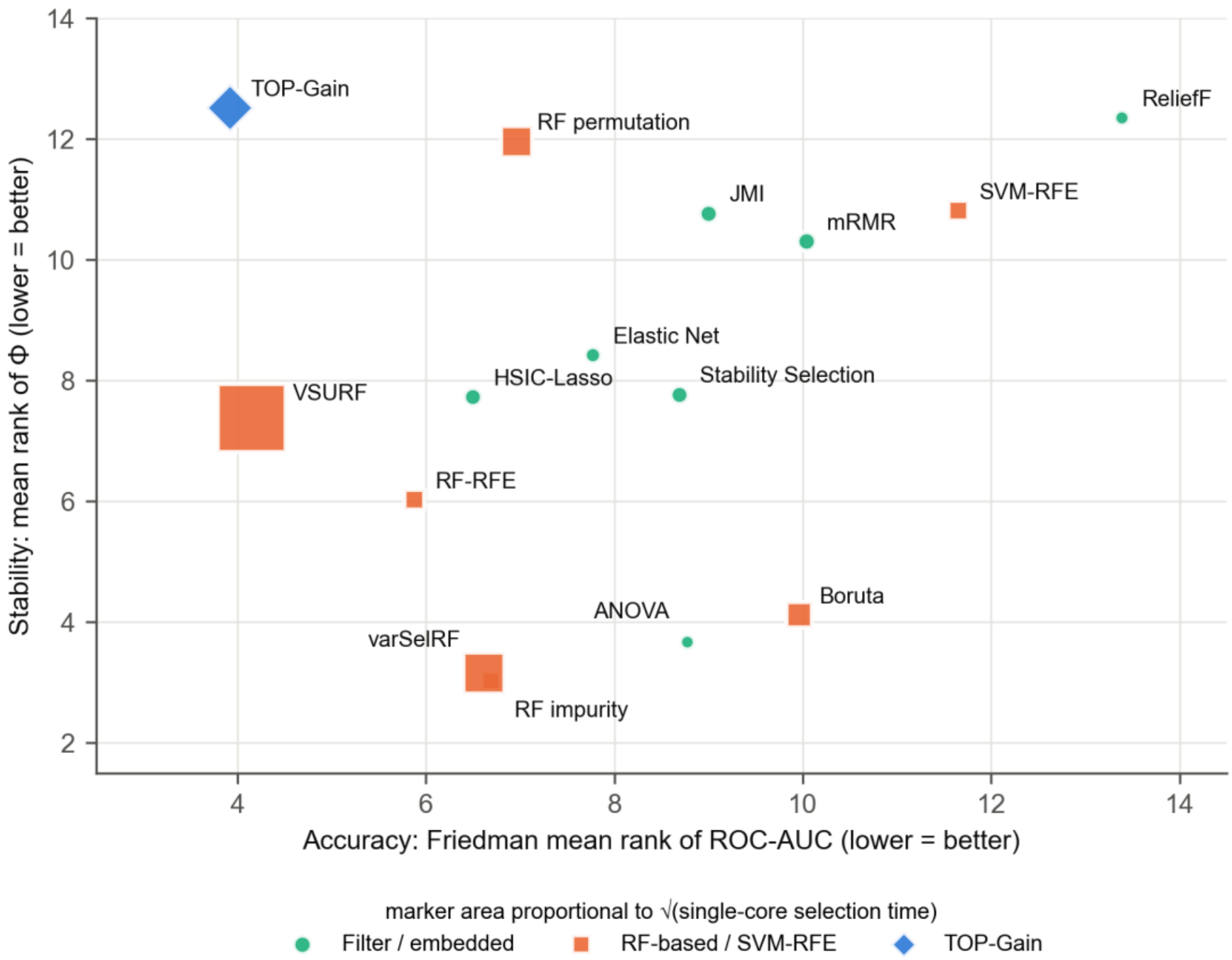

***Fig. 5*** *Accuracy–stability–cost trade-off on CC18. Horizontal axis: Friedman mean rank of ROC-AUC; vertical axis: mean rank of the stability Φ (lower is better on both axes). Marker area grows with the single-core selection time (Table 6)*

## 4.6 Why truncation helps: ranking reliability

To test the mechanism directly, we measured how reproducibly each criterion ranks the candidates at a fixed step of the search. We used six datasets (synthetic, qsar-biodeg, MiceProtein, splice, Bioresponse and OVA_Breast) and fixed $S$ to the first $s \in \{1,3,6,10\}$ features chosen by TOP-Gain on the training part of the first fold. All remaining candidates were then scored 20 times under two kinds of perturbation: changing only the forest seeds, or drawing stratified 80% subsamples of the training data. Over the 190 pairs of replicates, we computed the mean pairwise Spearman correlation of the candidate scores, the agreement of the best candidate and of the five best candidates, and the intraclass correlation (ICC). As a diagnostic, we also scored the negative part that TOP-Gain discards.

**Table 7** Reproducibility of the candidate ranking (mean over 6 datasets × 4 subset sizes; higher is better). † TOP-Gain significantly higher than every other criterion (Wilcoxon over 24 units, α = 0.05).

| Criterion | Spearman (seed) | Spearman (data) | Top-5 (seed) | Top-5 (data) | Top-1 (seed) | Top-1 (data) | ICC (data) |
|---|---|---|---|---|---|---|---|
| **TOP-Gain** | **0.617†** | **0.606†** | **0.509** | **0.423†** | 0.454 | 0.358 | **0.547†** |
| Net probability gain | 0.511 | 0.538 | 0.446 | 0.383 | 0.443 | **0.411** | 0.391 |
| OOB Brier score | 0.539 | 0.514 | 0.471 | 0.382 | **0.497** | 0.408 | 0.391 |
| OOB log-loss | 0.520 | 0.492 | 0.438 | 0.373 | 0.459 | 0.387 | 0.377 |
| OOB accuracy | 0.428 | 0.356 | 0.356 | 0.279 | 0.378 | 0.301 | 0.336 |
| OOB balanced accuracy | 0.417 | 0.344 | 0.348 | 0.281 | 0.388 | 0.306 | 0.330 |
| *Negative part (discarded)* | *0.434* | *0.486* | *0.257* | *0.250* | *0.227* | *0.220* | *0.424* |

TOP-Gain ranks the candidates most reproducibly under both perturbations (Table 7), with Spearman correlations of 0.62 and 0.61 against 0.34–0.54 for the alternatives ($p \le 0.0002$). Under data perturbation, which corresponds to the fold-to-fold variation measured by $\Phi$, it also has the most reproducible set of five leading candidates and the largest share of between-candidate variance (ICC 0.55 versus 0.33–0.39). Its agreement on the single best candidate, however, is not higher. The stability gain therefore does not come from a more reproducible winner at each step, but from a more reproducible group of leading candidates: even when the winner changes, the features added over successive steps overlap more, and this is what the set-level measure Φ rewards.

The discarded negative part is the least reproducible component of all, with top-1 and top-5 agreement of only 0.22–0.26. Subtracting it, as the net gain does, adds a mostly noisy term to an informative one. The advantage of TOP-Gain is largest in the middle of the search, where many candidates offer similar gains: under data perturbation its Spearman correlation is 0.79 versus 0.68 for the best alternative at $|S| = 3$, and 0.50 versus 0.40 at $|S| = 6$. This matches the moderate-noise regime of the toy simulation (Fig. 2).

## 4.7 Ground truth

Synthetic data with a known ground truth confirm that the more stable selections are also more correct. With ten informative features among 50, TOP-Gain recovered 10.0, 10.0, 9.8 and 9.4 informative features at class-imbalance ratios ρ = 1, 5, 10 and 20, whereas the net gain recovered 9.6, 9.0, 9.0 and 7.8; TOP-Gain also had the highest stability at every ratio ($\Phi \ge 0.85$; Online Resource 1, Table S3a). At the most extreme ratio, ρ = 20, the proper scoring rules nevertheless gave higher minority recall and Macro-F1 than TOP-Gain (Macro-F1 0.742 for Brier versus 0.713), although they recovered fewer informative features. In an XOR benchmark, TOP-Gain recovered the most informative features for $K \le 4$, but, like every greedy forward criterion, it did not reliably recover the pure interaction pair.

## 4.8 Sensitivity to the surrogate configuration

Table 8 lists the variants of the surrogate configuration and their effects relative to the default.

**Table 8** Sensitivity of TOP-Gain to its surrogate configuration. Each variant changes one setting of the default ($T = 40$, depth 5, pool factor 1). Δ is the variant minus default TOP-Gain over datasets (W/L = wins and losses of the variant). $p_{\text{Holm}}$ is the Wilcoxon $p$ corrected over the variants. "AUC rank" is the variant's Friedman rank of ROC-AUC when it replaces default TOP-Gain among the 15 CC18 methods. Time is the median selection time per fold (ten cores).

| Variant | ΔAUC (W/L, $p_{\text{Holm}}$) | ΔMacro-F1 ($p_{\text{Holm}}$) | ΔΦ [95% CI] (W/L, $p_{\text{Holm}}$) | AUC rank (position) | Time/fold (s) |
|---|---|---|---|---|---|
| Default ($T = 40$, depth 5) | – | – | – | 3.92 (1st) | 4.9 |
| $T = 20$ | −0.001 (8/18, 0.15) | −0.002 (0.20) | −0.061 [−0.078, −0.044] (2/24, <0.001) | 4.38 (2nd) | 2.2 |
| $T = 80$ | +0.001 (17/9, 0.15) | +0.000 (0.97) | +0.052 [0.040, 0.067] (26/0, <0.001) | 3.58 (1st) | 7.1 |
| $T = 160$ | +0.001 (18/8, 0.12) | +0.001 (0.43) | +0.105 [0.086, 0.124] (25/1, <0.001) | 3.62 (1st) | 13.5 |
| depth 3 | −0.002 (6/20, 0.051) | −0.004 (0.001) | +0.038 [0.013, 0.064] (18/8, 0.009) | 4.62 (2nd) | 3.6 |
| depth 8 | +0.001 (16/10, 0.18) | +0.001 (0.97) | −0.036 [−0.059, −0.015] (7/19, 0.009) | 3.42 (1st) | 4.2 |
| pool × 0.5 (AMLB, $P > 150$) | −0.001 (6/10, 0.87) | −0.003 (0.70) | +0.005 [−0.018, 0.027] (9/7, 0.96) | – | 13.9 (default 27.2) |

| Variant | ΔAUC (W/L, $p_{\text{Holm}}$) | ΔMacro-F1 ($p_{\text{Holm}}$) | ΔΦ [95% CI] (W/L, $p_{\text{Holm}}$) | AUC rank (position) | Time/fold (s) |
|---|---|---|---|---|---|
| pool × 2 (AMLB, $P > 150$) | −0.002 (4/9, 0.87) | −0.001 (0.86) | −0.002 [−0.016, 0.015] (4/8, 0.96) | – | 39.7 (default 27.2) |

Predictive performance is insensitive to the surrogate configuration. Every change of $T$ between 20 and 160 or of the depth between 3 and 8 moves the mean ROC-AUC by at most 0.0023, and no ROC-AUC difference is significant after Holm correction. With every variant TOP-Gain remains first or second among the 15 CC18 methods, and the only method ever ahead of it is VSURF; only the shallowest trees (depth 3) cost slightly more accuracy and Macro-F1 than the other settings.

Stability, in contrast, depends on $T$, as the noise mechanism of Section 2.3 predicts, since more trees give each instance more OOB trees and thus less estimation noise. Relative to $T = 40$, Φ changes by −0.061 at $T = 20$, +0.052 at $T = 80$ and +0.105 at $T = 160$. At $T = 160$, TOP-Gain moves from the 15th to the 10th stability rank among the 15 methods, at 2.8 times the selection time of the default. Deeper trees (depth 8) reduce stability and shallower trees (depth 3) increase it, while the candidate-pool size has no detectable effect on accuracy or stability, so halving the pool halves the cost at no loss.

These results show that $T = 40$ is a reasonable compromise between cost and accuracy that does not favour TOP-Gain's accuracy results, and that the number of trees is a direct handle on stability that users can turn up when the reproducibility of the selected set matters more than selection time. Table 8 compares the variants only with default TOP-Gain, because the net-gain criterion was run only at $T = 40$ and a comparison with it would not isolate the effect of truncation at other values of $T$.

# 5 Discussion

The results support one consistent account of what truncation does. At a typical forward step, most candidates change the OOB probabilities of most instances only by noise. Criteria that average these changes, whether the net gain, the Brier score or the log-loss, let this noise reorder near-equivalent candidates from one fold to the next. Truncation removes most of that noise variance while keeping the instances that a candidate truly improves (Section 2.3). The effect is visible at three levels: in the reproducibility of the candidate ranking (Section 4.6), in the stability of the selected sets on 50 datasets from two independent suites (Sections 4.2 and 4.3), and in a confirmed moderator, as the gain is largest when relevant features are sparse and most candidates are therefore noise (H4b). Truncation does not cost accuracy: TOP-Gain was never less accurate than the net gain and was equal to or better than the proper scoring rules.

The noise argument of Section 2.3 rests on the assumption that negative changes in the OOB probabilities are mostly estimation noise, and TOP-Gain indeed does not penalise real harm. Suppose candidate A improves every instance by 0.02, whereas candidate B improves 30% of the instances by 0.10 and worsens the other 70% by 0.10. The net gain is 0.020 for A and −0.040 for B, but TOP-Gain is 0.020 for A and 0.030 for B, so TOP-Gain prefers the harmful B. Our results suggest that such cases are rare in practice. The negative part that TOP-Gain discards was the least reproducible component of the score (Section 4.6), which indicates that it consists mostly of noise rather than harm; a Random Forest makes little use of a feature in the regions where it does not help, so large systematic harm from a single added feature is uncommon; and TOP-Gain was never less accurate than the net gain on any of the two benchmark suites (Sections 4.2, 4.3). The assumption can nevertheless fail under severe class imbalance, where real harm to the few minority instances is outweighed by small gains on the majority class. At an imbalance ratio of 20, the proper scoring rules indeed gave higher minority recall and Macro-F1 than TOP-Gain, although TOP-Gain recovered more of the informative features (Section 4.7). A class-weighted variant of TOP-Gain, which would give the minority instances more weight in the truncated sum, is a possible remedy that we have not tested.

The RF-OOB forward-selection framework itself is not new [22]; our contribution is the criterion and the evidence for it. In practical terms, TOP-Gain reaches the accuracy of the best RF wrapper we tested, VSURF, at a small fraction of its cost, and it is significantly more accurate than the 13 other baselines on CC18 and all 12 baselines on AMLB. Its accuracy advantage is broad rather than tied to an identifiable type of dataset, since neither the pre-registered H1–H3 nor H4a was confirmed.

TOP-Gain is therefore a good choice when the predictive performance of a small panel is the main goal and selection times of seconds to minutes per fold are acceptable. When the exact identity of the selected features matters more than accuracy, more stable but less accurate rankings, such as RF impurity, ANOVA or Boruta, may be preferable. TOP-Gain's absolute stability is low on CC18 and intermediate on AMLB, but it can be raised without loss of accuracy by a larger surrogate forest; with $T = 160$, $\Phi$ increases by 0.10 at 2.8 times the cost (Section 4.8). Combining TOP-Gain with subsample aggregation, as in Stability Selection [12] or randomised forward selection [35], is a natural further step and is left for future work.

The study has several limitations. We examined the criterion only within greedy forward selection; floating or beam search might change the absolute numbers, and although the criterion does not depend on the search, we did not test other search strategies. As discussed above, TOP-Gain ignores real harm, which can matter under severe class imbalance. TOP-Gain is the most stable wrapper criterion but not the most stable selector (Sections 4.1 and 4.3), and, like all greedy forward methods, it cannot detect features that are informative only jointly until one of them has been selected. The scope of the comparisons is also limited in three respects: varSelRF and VSURF were run on CC18 only, so the comparison with VSURF rests on 26 datasets; the absolute stability values hold for the default $T = 40$, because stability depends on $T$ (Section 4.8); and all datasets were subsampled to at most 1,500 instances. Finally, CC18 and AMLB contain few small-sample, very high-dimensional datasets. In our original benchmark, which includes microarray and proteomics data, TOP-Gain also ranked first, but with smaller and less significant margins (Online Resource 1, Section S2).

# 6 Conclusion

TOP-Gain scores a candidate feature by the truncated sample-level gain in the out-of-bag Random Forest probability of the true class, and in doing so it removes most of the estimation noise that other continuous criteria accumulate. In a pre-registered benchmark on 26 CC18 datasets and a confirmatory study on 24 AutoML Benchmark datasets, TOP-Gain achieved the best predictive rank and was significantly better than 13 of 14 and 12 of 12 baselines, respectively, while matching the accuracy of VSURF at a fraction of its cost. Within an identical wrapper, it gave the most stable selections of six criteria at equal or better accuracy, and, as hypothesised in advance, its stability gain was largest when relevant features were sparse. Because it requires only a one-line change, we recommend TOP-Gain as the default criterion for Random-Forest forward wrapper selection.

# Declarations

**Funding.** This research was supported by a research fund from the Dankook University in 2025

**Competing interests.** The authors declare no competing interests.

**Data availability.** All datasets are publicly available on OpenML; their dataset IDs are listed in Online Resource 1 (Table S1). The per-fold results, the screening data and the protocol with all amendments are available in the code repository.

**Code availability.** The complete code is available at https://github.com/dkumango/TOP_Gain. It covers the selectors, benchmark runners, analysis scripts and the pre-registration protocol. The main scripts are:

- `topg_dataset_screening.py`: eligibility rules and moderators;
- `topg_run_cc18.py`: CC18 and AMLB benchmarks;
- `topg_analyze_cc18.py` and `topg_analyze_amlb.py`: pre-registered analyses;
- `topg_runtime_benchmark.py`: runtime study;
- `topg_run_sensitivity.py`: sensitivity study.

**Ethics approval and consent.** Not applicable. The study uses only public, de-identified benchmark datasets.

**Author contributions.** Oh: Conceptualization, Methodology, Funding acquisition, Investigation, Project administration, Validation, Writing – review and editing.

**Use of generative AI.** The use of an AI assistant is described in Section 3.7.

# Online Resource 1

**Truncated Out-of-Bag Probability Gain: A Noise-Robust Criterion for Random-Forest Wrapper Feature Selection**

*Knowledge and Information Systems*

[Author names] · Corresponding author: [Name], [Affiliation, City, Country], [e-mail]

This Online Resource contains:

- S1 the dataset lists;
- S2 the original ten-dataset benchmark;
- S3 the ground-truth and controlled-imbalance studies;
- S4 the robustness and ablation checks;
- S5 the results per downstream classifier;
- S6 the protocol amendments.

## S1 Datasets

Tables S1a and S1b list the eligible datasets of the two suites with their moderators, sorted by the number of features.

**Table S1a** CC18 datasets (26 eligible; sorted by $P$). ID = OpenML dataset ID; $N$ = instances in the original data; $N_{\text{used}}$ = instances used after subsampling to at most 1,500; $C$ = classes; Imb. = majority/minority ratio; moderators as defined in Section 3.1 of the main text.

| Dataset | ID | $N$ | $N_{\text{used}}$ | $P$ | $C$ | Imb. | M-rel | M-nl | M-red | M-head |
|---|---|---|---|---|---|---|---|---|---|---|
| climate-model-simulation-crashes | 40994 | 540 | 540 | 18 | 2 | 10.7 | 0.22 | -0.05 | 0.02 | 0.10 |
| pc1 | 1068 | 1,109 | 1,109 | 21 | 2 | 13.4 | 0.95 | 0.02 | 0.90 | 0.15 |
| kc2 | 1063 | 522 | 522 | 21 | 2 | 3.9 | 1.00 | -0.02 | 0.91 | 0.19 |
| kc1 | 1067 | 2,109 | 1,500 | 21 | 2 | 5.5 | 0.95 | 0.02 | 0.90 | 0.17 |
| jm1 | 1053 | 10,885 | 1,500 | 21 | 2 | 4.2 | 1.00 | -0.02 | 0.89 | 0.33 |
| steel-plates-fault | 40982 | 1,941 | 1,500 | 27 | 7 | 12.4 | 1.00 | 0.03 | 0.83 | 0.04 |
| churn | 40701 | 5,000 | 1,500 | 29 | 2 | 6.1 | 0.46 | 0.06 | 0.47 | 0.08 |
| sick | 38 | 3,772 | 1,500 | 30 | 2 | 15.3 | 0.38 | 0.03 | 0.41 | 0.01 |
| wdbc | 1510 | 569 | 569 | 30 | 2 | 1.7 | 0.83 | -0.01 | 0.85 | 0.01 |
| GesturePhaseSegmentationProcessed | 4538 | 9,873 | 1,500 | 32 | 5 | 2.9 | 0.56 | 0.07 | 0.75 | 0.18 |
| satimage | 182 | 6,430 | 1,500 | 36 | 6 | 2.4 | 1.00 | 0.01 | 0.94 | 0.02 |
| pc3 | 1050 | 1,563 | 1,500 | 37 | 2 | 8.7 | 0.86 | 0.03 | 0.90 | 0.16 |
| pc4 | 1049 | 1,458 | 1,458 | 37 | 2 | 7.2 | 0.84 | 0.04 | 0.86 | 0.06 |
| PhishingWebsites | 4534 | 11,055 | 1,500 | 38 | 2 | 1.3 | 0.71 | -0.00 | 0.55 | 0.02 |
| qsar-biodeg | 1494 | 1,055 | 1,055 | 41 | 2 | 2.0 | 0.83 | 0.01 | 0.70 | 0.06 |
| mfeat-zernike | 22 | 2,000 | 1,500 | 47 | 10 | 1.0 | 1.00 | -0.01 | 0.87 | 0.03 |
| credit-g | 31 | 1,000 | 1,000 | 48 | 2 | 2.3 | 0.50 | 0.00 | 0.40 | 0.21 |
| first-order-theorem-proving | 1475 | 6,118 | 1,500 | 51 | 6 | 5.3 | 0.78 | 0.07 | 0.78 | 0.22 |
| spambase | 44 | 4,601 | 1,500 | 57 | 2 | 1.5 | 0.91 | 0.01 | 0.46 | 0.02 |
| ozone-level-8hr | 1487 | 2,534 | 1,500 | 72 | 2 | 14.8 | 0.89 | 0.02 | 0.88 | 0.10 |
| mfeat-fourier | 14 | 2,000 | 1,500 | 76 | 10 | 1.0 | 1.00 | 0.01 | 0.43 | 0.02 |
| cylinder-bands | 6332 | 540 | 540 | 143 | 2 | 1.4 | 0.15 | 0.06 | 0.35 | 0.10 |
| nomao | 1486 | 34,465 | 1,500 | 145 | 2 | 2.5 | 0.87 | 0.00 | 0.95 | 0.02 |
| madelon | 1485 | 2,600 | 1,500 | 500 | 2 | 1.0 | 0.03 | 0.19 | 0.12 | 0.26 |

| Dataset | ID | $N$ | $N_{used}$ | $P$ | $C$ | Imb. | M-rel | M-nl | M-red | M-head |
|---|---|---|---|---|---|---|---|---|---|---|
| Internet-Advertisements | 40978 | 3,279 | 1,500 | 1,558 | 2 | 6.1 | 0.25 | 0.00 | 0.92 | 0.02 |
| Bioresponse | 4134 | 3,751 | 1,500 | 1,776 | 2 | 1.2 | 0.24 | 0.10 | 0.68 | 0.16 |

**Table S1b** AMLB datasets (24 eligible; OpenML suite 271 minus CC18; sorted by $P$).

| Dataset | ID | $N$ | $N_{used}$ | $P$ | $C$ | Imb. | M-rel | M-nl | M-red | M-head |
|---|---|---|---|---|---|---|---|---|---|---|
| sylvine | 41146 | 5,124 | 1,500 | 20 | 2 | 1.0 | 0.30 | 0.01 | 0.33 | 0.03 |
| Higgs | 42769 | 1,000,000 | 1,500 | 28 | 2 | 1.1 | 0.21 | 0.03 | 0.33 | 0.29 |
| Satellite | 40900 | 5,100 | 1,500 | 36 | 2 | 67.2 | 0.75 | -0.02 | 0.94 | 0.04 |
| ada | 41156 | 4,147 | 1,500 | 48 | 2 | 3.0 | 0.67 | 0.00 | 0.34 | 0.10 |
| MiniBooNE | 41150 | 130,064 | 1,500 | 50 | 2 | 2.6 | 0.10 | 0.05 | 0.68 | 0.04 |
| covertype | 1596 | 581,012 | 1,500 | 54 | 7 | 104.4 | 0.71 | 0.01 | 0.30 | 0.09 |
| jannis | 41168 | 83,733 | 1,500 | 54 | 4 | 23.0 | 0.46 | 0.03 | 0.42 | 0.21 |
| jasmine | 41143 | 2,984 | 1,500 | 144 | 2 | 1.0 | 0.40 | 0.04 | 0.38 | 0.14 |
| APSFailure | 41138 | 76,000 | 1,500 | 170 | 2 | 54.6 | 0.70 | 0.07 | 0.82 | 0.02 |
| volkert | 41166 | 58,310 | 1,500 | 180 | 10 | 9.4 | 0.95 | 0.04 | 0.84 | 0.13 |
| porto-seguro | 42742 | 595,212 | 1,500 | 204 | 2 | 26.3 | 0.02 | 0.13 | 0.29 | 0.35 |
| madeline | 41144 | 3,140 | 1,500 | 259 | 2 | 1.0 | 0.04 | 0.23 | 0.14 | 0.22 |
| philippine | 41145 | 5,832 | 1,500 | 308 | 2 | 1.0 | 0.39 | 0.04 | 0.51 | 0.20 |
| fabert | 41164 | 8,237 | 1,500 | 800 | 7 | 3.9 | 0.22 | 0.03 | 0.37 | 0.15 |
| gina | 41158 | 3,153 | 1,500 | 970 | 2 | 1.0 | 0.38 | 0.07 | 0.81 | 0.03 |
| christine | 41142 | 5,418 | 1,500 | 1,611 | 2 | 1.0 | 0.40 | 0.10 | 0.49 | 0.22 |
| dilbert | 41163 | 10,000 | 1,500 | 2,000 | 5 | 1.1 | 0.84 | -0.00 | 0.82 | 0.02 |
| kick | 41162 | 72,983 | 1,500 | 2,408 | 2 | 7.2 | 0.02 | 0.07 | 0.61 | 0.36 |
| riccardo | 41161 | 20,000 | 1,500 | 4,296 | 2 | 3.0 | 0.02 | 0.07 | 0.45 | 0.07 |
| guillermo | 41159 | 20,000 | 1,500 | 4,296 | 2 | 1.5 | 0.01 | -0.01 | 0.47 | 0.36 |
| robert | 41165 | 10,000 | 1,500 | 7,200 | 10 | 1.1 | 0.90 | 0.04 | 0.86 | 0.21 |
| amazon-commerce-reviews | 1457 | 1,500 | 1,500 | 10,000 | 50 | 1.0 | 0.70 | -0.04 | 0.56 | 0.05 |
| arcene | 41157 | 100 | 100 | 10,000 | 2 | 1.3 | 0.08 | -0.04 | 0.81 | 0.16 |
| KDDCup09_appetency | 1111 | 50,000 | 1,500 | 71,640 | 2 | 54.6 | 0.02 | 0.09 | 0.93 | 0.34 |

## S2 Original ten-dataset benchmark, including P ≫ N data

**Datasets.** The original benchmark used ten biomedical and life-science datasets from OpenML:

- a synthetic benchmark;
- mouse protein expression;
- three tumour-expression microarray tasks;
- ovarian-cancer proteomics;
- three molecular or sequence datasets.

These datasets have up to 15,154 features and include microarray and proteomics data with $N \ll P$.

**Design.**

- *Selectors.* TOP-Gain was compared with nine baselines (ANOVA, mRMR, JMI, ReliefF, Elastic Net, HSIC-Lasso, Stability Selection, Boruta, SVM-RFE) and with the five alternative criteria.
- *Panel sizes.* For $P \leq 500$, $K = \max(2, \lfloor rP \rfloor)$ with the fraction $r \in \{0.1, 0.2, 0.3, 0.4\}$ of the features. For $P > 500$, $K \in \{10, 20, 30, 50\}$.

- *Protocol.* The four downstream classifiers and the statistics are those of the main text. Five repeats of 5-fold cross-validation were run, with seeds 42, 1, 2, 3 and 4.

**Results (Table S2a and Table S2b).**

- TOP-Gain had the best Friedman mean rank for ROC-AUC (2.90), accuracy (2.80) and Macro-F1 (2.90) among ten methods (Friedman $p \leq 0.018$).
- Its mean ROC-AUC exceeded that of every baseline by 0.006–0.028, with bootstrap CIs excluding zero. With ten datasets, however, the difference was significant after Holm correction only against SVM-RFE and ReliefF.
- All six criteria had equal accuracy (Friedman $p \geq 0.39$). TOP-Gain was the most stable criterion (mean rank 1.40) and significantly more stable than Brier, log-loss, accuracy and balanced accuracy ($p_{\mathrm{Holm}} \leq 0.027$). It was more stable than the net gain on 7 of 10 datasets ($\Delta\Phi$ +0.026, $p_{\mathrm{Holm}} = 0.11$).

**Table S2a** Original benchmark, TOP-Gain vs nine baselines (10 datasets × 5 repeats; Friedman ranks among 10 methods, CD = 4.28).

| Method | Rank AUC ↓ | Rank Φ ↓ | ΔAUC [95% CI] (W/L, $p_{\mathrm{Holm}}$) | ΔΦ [95% CI] (W/L, $p_{\mathrm{Holm}}$) |
|---|---|---|---|---|
| **TOP-Gain** | 2.90 | 7.90 | – | – |
| ANOVA | 6.50 | 1.50 | +0.010 [0.003, 0.019] (7/3, 0.16) | -0.426 [-0.511, -0.341] (0/10, 0.02) |
| mRMR | 5.40 | 7.40 | +0.018 [0.001, 0.048] (7/3, 0.17) | -0.040 [-0.216, 0.145] (4/6, 1.00) |
| JMI | 5.20 | 6.00 | +0.013 [0.001, 0.032] (8/2, 0.16) | -0.113 [-0.296, 0.076] (5/5, 1.00) |
| ReliefF | 7.40 | 6.70 | +0.020 [0.004, 0.045] (9/1, 0.03) | -0.077 [-0.236, 0.078] (4/6, 1.00) |
| ElasticNet | 5.60 | 4.60 | +0.007 [0.002, 0.014] (8/2, 0.16) | -0.201 [-0.301, -0.109] (0/10, 0.02) |
| HSIC-Lasso | 3.60 | 4.90 | +0.006 [0.001, 0.014] (7/3, 0.17) | -0.175 [-0.304, -0.041] (2/8, 0.24) |
| Stability Selection | 4.90 | 5.30 | +0.009 [0.002, 0.017] (7/3, 0.16) | -0.165 [-0.237, -0.101] (0/10, 0.02) |
| Boruta | 5.90 | 2.70 | +0.022 [0.006, 0.043] (8/2, 0.10) | -0.336 [-0.389, -0.284] (0/10, 0.02) |
| SVM-RFE | 7.60 | 8.00 | +0.028 [0.009, 0.050] (10/0, 0.02) | +0.028 [-0.063, 0.121] (6/4, 1.00) |

**Table S2b** Original benchmark, criterion comparison (Friedman ranks among 6 criteria, CD = 2.38).

| Method | Rank AUC ↓ | Rank Φ ↓ | ΔAUC [95% CI] (W/L, $p_{\mathrm{Holm}}$) | ΔΦ [95% CI] (W/L, $p_{\mathrm{Holm}}$) |
|---|---|---|---|---|
| **TOP-Gain** | 3.90 | 1.40 | – | – |
| Net probability gain | 2.70 | 2.30 | -0.001 [-0.002, 0.000] (4/6, 1.00) | +0.025 [-0.002, 0.054] (7/3, 0.11) |
| OOB Brier score | 3.00 | 3.60 | -0.000 [-0.002, 0.002] (3/7, 1.00) | +0.074 [0.044, 0.105] (10/0, 0.01) |
| OOB log-loss | 3.30 | 3.40 | -0.000 [-0.002, 0.001] (4/6, 1.00) | +0.074 [0.029, 0.120] (9/1, 0.03) |
| OOB accuracy | 3.80 | 5.00 | +0.001 [-0.001, 0.002] (5/5, 1.00) | +0.146 [0.108, 0.184] (10/0, 0.01) |
| OOB balanced accuracy | 4.30 | 5.30 | +0.001 [-0.001, 0.002] (5/5, 1.00) | +0.161 [0.126, 0.203] (10/0, 0.01) |

# S3 Ground-truth and controlled-imbalance studies

**Synthetic imbalance sweep.**

- *Data.* $N = 2{,}000$, $P = 50$. Ten features are informative (a mean shift of the minority class) and 40 are noise.
- *Design.* Imbalance ratios $\rho \in \{1,5,10,20\}$ and $K = 10$. The downstream classifier is a class-balanced RF.
- *Measures.* Minority recall, balanced accuracy, Macro-F1, ROC-AUC, Φ, and the number of informative features selected (TP).

**Table S3a** Synthetic imbalance sweep ($K = 10$; 10 informative features among 50). TP = number of informative features selected (mean over five folds).

| $\rho$ | Criterion | Minority recall | Bal. acc. | Macro-F1 | ROC-AUC | TP (/10) | Φ |
|---|---|---|---|---|---|---|---|
| 1 | **TOP-Gain** | 0.914 | **0.922** | **0.922** | **0.980** | **10.0** | **1.000** |
| 1 | Net gain | 0.906 | 0.918 | 0.917 | 0.977 | 9.6 | 0.900 |
| 1 | OOB Brier | **0.917** | 0.920 | 0.920 | 0.977 | 9.8 | 0.950 |
| 1 | OOB accuracy | 0.911 | 0.921 | 0.921 | 0.976 | 9.2 | 0.838 |
| 5 | **TOP-Gain** | **0.691** | **0.841** | **0.880** | **0.984** | **10.0** | **1.000** |
| 5 | Net gain | **0.691** | 0.839 | 0.876 | 0.976 | 9.0 | 0.788 |
| 5 | OOB Brier | **0.691** | 0.840 | 0.878 | 0.975 | 9.0 | 0.763 |
| 5 | OOB accuracy | 0.670 | 0.830 | 0.872 | 0.975 | 8.6 | 0.663 |
| 10 | **TOP-Gain** | 0.525 | 0.761 | 0.826 | 0.976 | **9.8** | **0.950** |
| 10 | Net gain | **0.531** | **0.764** | **0.828** | **0.980** | 9.0 | 0.838 |
| 10 | OOB Brier | 0.520 | 0.759 | 0.820 | 0.977 | 9.0 | 0.838 |
| 10 | OOB accuracy | 0.448 | 0.720 | 0.775 | 0.956 | 7.0 | 0.388 |
| 20 | **TOP-Gain** | 0.305 | 0.652 | 0.713 | 0.962 | **9.4** | **0.850** |
| 20 | Net gain | 0.274 | 0.637 | 0.701 | 0.946 | 7.8 | 0.550 |
| 20 | OOB Brier | **0.347** | **0.673** | **0.742** | 0.955 | 8.4 | 0.688 |
| 20 | OOB log-loss | 0.326 | 0.662 | 0.729 | **0.965** | 8.4 | 0.625 |
| 20 | OOB accuracy | 0.253 | 0.625 | 0.680 | 0.938 | 6.6 | 0.400 |

**Results of the imbalance sweep (Table S3a).**

- TOP-Gain selected the most informative features at every imbalance ratio. It recovered all 10 at $\rho \leq 5$, and 9.8 and 9.4 at $\rho = 10$ and 20.
- It also had the highest stability throughout ($\Phi \geq 0.85$).
- At $\rho = 20$, the proper scoring rules gave higher minority recall and Macro-F1 than TOP-Gain, although they recovered fewer informative features. Selecting the right features does not by itself guarantee the best threshold-dependent metrics under severe imbalance.

**Epistasis benchmark.**

- *Data.* $N = 1{,}500$, $P = 30$. Two features ($F_1, F_2$) act only through an XOR interaction, four ($F_3$–$F_6$) have additive effects, four are noisy copies of the additive drivers, and 20 are pure noise.
- *Design.* $K \in \{2,4,6\}$.

**Table S3b** Epistasis benchmark (XOR pair $F_1, F_2$ with no marginal effect; additive drivers $F_3$–$F_6$). Values: informative features selected (/6) | RF ROC-AUC | Φ.

| Criterion | $K = 2$ | $K = 4$ | $K = 6$ |
|---|---|---|---|
| **TOP-Gain** | **1.8** \| 0.616 \| **0.786** | **3.6** \| **0.705** \| **0.769** | 3.8 \| 0.698 \| **0.646** |
| Net gain | 1.4 \| 0.618 \| 0.679 | 2.4 \| 0.685 \| 0.481 | 3.2 \| 0.699 \| 0.604 |
| OOB Brier | 1.2 \| 0.613 \| 0.464 | 2.4 \| 0.692 \| 0.423 | 3.2 \| 0.711 \| 0.500 |
| OOB log-loss | 1.2 \| 0.612 \| 0.357 | 2.8 \| 0.691 \| 0.423 | 3.2 \| 0.707 \| 0.354 |
| OOB accuracy | 1.0 \| 0.606 \| 0.571 | 2.0 \| 0.673 \| 0.510 | 2.8 \| 0.729 \| 0.438 |
| TOP-Gain + all extensions | 1.6 \| **0.623** \| 0.679 | 3.2 \| 0.696 \| 0.596 | **4.0** \| **0.736** \| 0.562 |

**Results of the epistasis benchmark (Table S3b).**

- For small panels ($K \leq 4$), TOP-Gain recovered the most informative features with the highest stability.

- No criterion reliably captured the pure XOR pair: it was recovered in at most 1 of 5 folds. Greedy forward selection cannot detect a feature whose marginal contribution is zero until its partner has been selected.

**OVA_Lung controlled imbalance.**

- *Data.* $P = 10{,}935$, with lung tumours as the minority class. We used the natural data ($\rho = 11.3$) and subsamples with 70 minority instances and $\rho \in \{1, 5, 10, 20\}$, each drawn with three seeds.
- *Design.* $K = 10$, with a class-balanced RF as the downstream classifier.

**Table S3c** OVA_Lung imbalance sweep ($K = 10$; balanced accuracy / ROC-AUC; mean over folds × seeds). "Natural" uses all 1,545 samples ($\rho = 11.3$).

| Method | Natural | $\rho = 1$ | $\rho = 5$ | $\rho = 10$ | $\rho = 20$ |
|---|---|---|---|---|---|
| **TOP-Gain** | 0.901 / **0.963** | 0.926 / 0.948 | 0.923 / 0.945 | 0.911 / 0.940 | 0.897 / 0.944 |
| Net gain | 0.921 / 0.955 | **0.933** / 0.952 | 0.925 / 0.941 | 0.916 / 0.939 | 0.904 / 0.945 |
| OOB Brier | 0.913 / 0.944 | 0.926 / 0.955 | 0.924 / 0.938 | 0.906 / 0.947 | 0.887 / 0.938 |
| OOB accuracy | 0.905 / 0.958 | 0.919 / 0.945 | 0.922 / 0.941 | 0.888 / 0.954 | 0.895 / **0.947** |
| TOP-Gain + all extensions | 0.929 / 0.949 | 0.929 / 0.953 | 0.925 / 0.951 | 0.916 / 0.944 | 0.909 / 0.943 |
| ANOVA | **0.929** / 0.934 | 0.931 / 0.935 | **0.929** / 0.941 | **0.921** / 0.940 | **0.916** / 0.935 |
| Elastic Net | **0.929** / 0.952 | 0.836 / 0.886 | 0.914 / 0.946 | 0.916 / 0.939 | 0.899 / 0.931 |
| mRMR | 0.905 / 0.959 | **0.933** / **0.958** | 0.916 / **0.962** | 0.896 / **0.955** | 0.861 / 0.946 |

**Results of the OVA_Lung study (Table S3c).**

- With ten-gene panels in a 10,935-gene space, the differences between methods were small and inconsistent across imbalance tiers. The ROC-AUC ranges of the wrapper criteria overlapped within about ±0.01.
- TOP-Gain had the highest ROC-AUC on the natural data, while ANOVA had the highest balanced accuracy in four of the five tiers.
- In this setting TOP-Gain's stability advantage over the net gain was not uniform (Table S3d).

**Table S3d** Stability Φ on OVA_Lung, TOP-Gain versus the net gain

| | Natural | $\rho = 1$ | $\rho = 5$ | $\rho = 10$ | $\rho = 20$ |
|---|---|---|---|---|---|
| TOP-Gain Φ | 0.25 | 0.26 | 0.28 | 0.26 | 0.29 |
| Net gain Φ | 0.39 | 0.21 | 0.23 | 0.28 | 0.30 |

# S4 Robustness and ablation checks

**Common-seed reference forest (Table S4a).** Algorithm 1 refits the reference probabilities with seed $s + \text{step}$. Refitting them with the candidates' seed $s$ gave equivalent results.

- ROC-AUC: $p = 0.36$; Φ: $p = 0.70$, original versus common seed at the dataset level.
- The Grand Rank in Table S4a is the mean rank over all evaluation slices: datasets × $K$ × classifiers × metrics.

**Table S4a** TOP-Gain with a common-seed reference forest

| Comparison | TOP-Gain (original) | TOP-Gain (common seed) |
|---|---|---|
| Mean ROC-AUC / Macro-F1 | 0.9519 / 0.9046 | 0.9514 / 0.9041 |
| Mean stability Φ | 0.480 | 0.485 |
| Original vs common seed (dataset level) | – | ROC-AUC $p = 0.36$; Φ $p = 0.70$ |

| Comparison | TOP-Gain (original) | TOP-Gain (common seed) |
|---|---|---|
| Grand Rank among TOP-Gain + nine baselines | 1st (3.49) | 1st (3.61) |
| Friedman mean rank of ROC-AUC (ten methods) | 1st (1.85) | 1st (2.55) |
| ΔROC-AUC vs baselines: 95% CI excludes zero | 9 of 9 | 9 of 9 |
| ROC-AUC vs baselines: significant after Holm | 5 of 9 | 2 of 9 (ReliefF, SVM-RFE) |
| Φ vs alternative criteria: significant after Holm | 5 of 5 ($p_{\text{Holm}} \leq 0.020$) | 5 of 5 ($p_{\text{Holm}} \leq 0.010$) |

**Multi-objective extensions (Table S4b).**

- *What was added.* A loss penalty (the weighted negative part), inverse-prevalence class weights, Laplacian smoothing of the OOB probabilities, and a pairwise synergy search among the ten best candidates.
- *Effect.* None of these improved predictive performance (all $p \geq 0.16$). They reduced stability on all ten datasets ($p = 0.002$) and increased runtime by 20–33%.
- *Synthetic imbalance sweep.* The fully extended variant was worse than plain TOP-Gain at every ratio (Table S4c).

**Table S4b** TOP-Gain versus TOP-Gain with extensions (ten-dataset benchmark; Grand Rank among these three methods).

| Variant | Grand Rank ↓ | ROC-AUC | Accuracy | Macro-F1 | Stability Φ | Time/fold (s) |
|---|---|---|---|---|---|---|
| **TOP-Gain** | **1.92** | 0.9519 | **0.9158** | **0.9046** | **0.480** | **66.1** |
| + loss penalty, class weights, smoothing | 1.98 | 0.9520 | 0.9150 | 0.9036 | 0.415 | 79.6 |
| + loss penalty, class weights, smoothing, pair synergy search | 2.10 | 0.9520 | 0.9145 | 0.9029 | 0.425 | 88.1 |

**Table S4c** Synthetic imbalance sweep, plain TOP-Gain versus the fully extended variant (Macro-F1 / informative features selected)

| $\rho$ | TOP-Gain Macro-F1 / TP | Fully extended Macro-F1 / TP |
|---|---|---|
| 5 | 0.880 / 10.0 | 0.870 / 9.0 |
| 20 | 0.713 / 9.4 | 0.704 / 7.6 |

# S5 Results per downstream classifier

Table S5 repeats the Friedman comparison of the main text separately for each downstream classifier.

**Table S5** Friedman mean rank of TOP-Gain among the baselines, per downstream classifier (position in parentheses; best other method and its rank).

| Classifier | CC18 ROC-AUC | CC18 Accuracy | CC18 Macro-F1 | AMLB ROC-AUC | AMLB Accuracy | AMLB Macro-F1 |
|---|---|---|---|---|---|---|
| Random Forest | 3.35 (1st; VSURF 4.46) | 3.60 (1st; VSURF 4.81) | 3.35 (1st; VSURF 4.31) | 3.00 (1st; RF-RFE 4.54) | 3.12 (1st; RF-RFE 4.08) | 3.54 (1st; RF-RFE 5.06) |
| XGBoost | 3.62 (1st; VSURF 4.62) | 4.29 (1st; VSURF 5.40) | 3.35 (1st; VSURF 4.88) | 2.38 (1st; RF perm. 4.75) | 3.27 (1st; RF-RFE 4.88) | 2.42 (1st; RF-RFE 4.54) |
| RBF-SVM | 4.85 (2nd; VSURF 4.81) | 4.04 (1st; VSURF 6.08) | 3.92 (1st; VSURF 5.35) | 3.00 (1st; RF-RFE 4.17) | 3.25 (1st; RF-RFE 4.29) | 3.79 (1st; RF-RFE 4.94) |
| Logistic regression | 5.04 (2nd; VSURF 4.69) | 4.58 (1st; VSURF 5.79) | 5.08 (1st; VSURF 5.69) | 3.29 (1st; RF-RFE 4.75) | 4.21 (1st; HSIC-Lasso 5.46) | 3.92 (1st; RF-RFE 5.54) |

# S6 Protocol amendments

The protocol was written before any selector was run on the benchmark data. Each change was appended as a dated amendment with its reason, and the full protocol is part of the code repository.

| Amendment | Date | Type | Content |
|---|---|---|---|
| A1 | 2026-10-11 | Implementation fix | Drop all-missing columns before imputation (dataset sick); re-screen. |
| A2 | 2026-10-11 | Design decision, before any selector run | Fixed panel sizes $K \in \{5,10,15,20\}$, $K \leq P/2$. |
| A3 | 2026-10-12 | Registration, before any AMLB result | H1–H3 not supported; H4 and the AMLB pool registered. |
| A4 | 2026-10-13 | Implementation fix, output unchanged | Memory-efficient loader (verified bit-identical); re-run policy for infrastructure errors. |
| A5 | 2026-10-14 | Efficiency fix, selections verified identical | RF permutation passes NumPy arrays. |
| A6 | 2026-10-15 | Crash fix, non-crashing runs unchanged | JMI choice when all scores are NaN. |
| A7 | 2026-10-20 | Re-runs under A4 | Four AMLB cells re-run once; all succeeded. |
| A8 | 2026-10-20 | Registration, before the study was run | Hyperparameter sensitivity study (T, depth, pool size); completed without failures. |